\documentclass{article}

\usepackage[square,numbers]{natbib}
\usepackage[preprint]{neurips_2023}
\usepackage{wrapfig}

\usepackage{microtype}
\usepackage{graphicx}
\usepackage{subfigure}
\usepackage{booktabs} %
\usepackage{enumitem}

\usepackage{hyperref}

\setenumerate[1]{align=left,label=\arabic*}
\usepackage{cancel}

\usepackage[ruled, vlined, linesnumbered,algo2e]{algorithm2e}
\SetKwInput{KwData}{Inputs}
\SetKwInput{KwResult}{Output}

\usepackage[utf8]{inputenc} %
\usepackage[T1]{fontenc}    %
\usepackage{url}            %
\usepackage{amsfonts}       %
\usepackage{nicefrac}       %
\usepackage{amsmath}
\usepackage{amssymb}
\usepackage{mathtools}
\usepackage{amsthm}
\usepackage{bm, bbm}

\usepackage{nth}

\usepackage{multirow}
\usepackage{float}
\usepackage{dblfloatfix}
\usepackage{algorithm2e}

\usepackage[capitalize,noabbrev]{cleveref}
\crefformat{equation}{(#2#1#3)}
\crefformat{figure}{Figure #2#1#3}
\crefformat{table}{Table #2#1#3}
\crefformat{appendix}{Appendix #2#1#3}
\crefformat{section}{Section #2#1#3}

\theoremstyle{plain}

\theoremstyle{definition}

\theoremstyle{remark}

\newcommand{\dpre}{d_{\mathrm{pre}}}
\newcommand{\dlev}{d_{\mathrm{lev}}}
\newcommand{\dsub}{d_{\mathrm{sub}}}
\newcommand{\dtheta}{d_{\theta}}

\newcommand{\R}{\mathbb{R}}

\usepackage[utf8]{inputenc} %
\usepackage[T1]{fontenc}    %
\usepackage{url}            %
\usepackage{hyperref}       %
\usepackage{booktabs}       %
\usepackage{amsfonts}       %
\usepackage{nicefrac}       %
\usepackage{microtype}      %
\usepackage{xcolor}         %

\title{Image Reconstruction via Deep Image Prior Subspaces
}

\author{%
  Riccardo Barbano \\
  University College London \\
  \And
  Javier Antorán \\
  University of Cambridge \\
  \And
  Johannes Leuschner \\
  University of Bremen \\
  \And
  José Miguel Hernández-Lobato \\
  University of Cambridge \\
  \And 
  Bangti Jin \\
  \shortstack{\phantom{}\\[-0.225em]The Chinese University\\[-0.2em]of Hong Kong}
  \And
  \v Zeljko Kereta \\
  University College London \\
}

\begin{document}
\maketitle

\begin{abstract}
Deep learning has been widely used for solving image reconstruction tasks but its deployability has been held back due to the shortage of high-quality training data.
Unsupervised learning methods, such as the deep image prior (DIP), naturally fill this gap, but bring a host of new issues: the susceptibility to overfitting due to a lack of robust early stopping strategies and unstable convergence.
We present a novel approach to tackle these issues by restricting DIP optimisation to a sparse linear subspace of its parameters, employing a synergy of dimensionality reduction techniques and second order optimisation methods.
The low-dimensionality of the subspace reduces DIP's tendency to fit noise and allows the use of stable second order optimisation methods, e.g., natural gradient descent or L-BFGS.
Experiments across both image restoration and tomographic tasks of different geometry and ill-posedness show that second order optimisation within a low-dimensional subspace is favourable in terms of optimisation stability to reconstruction fidelity trade-off.
\end{abstract}

\section{Introduction}\label{sec:intro}

Deep learning (DL) approaches have shown impressive results in a wide variety of linear inverse problems in imaging, e.g., denoising \citep{Tian2020denoising}, super-resolution \citep{ledig2017super,ulyanov2020dip}, magnetic resonance imaging \citep{Zeng2021mrireview} and tomographic reconstruction \citep{wang2020deep}. 
Mathematically, a linear inverse problem is formulated as the recovery of an unknown image ${x} \in \mathbb{R}^{d_{x}}$ from measurements $y \in \mathbb{R}^{d_{y}}$ given by
\begin{equation}\label{eq:inverse_problem}
    y = A x + \epsilon,
\end{equation}
for an (ill-conditioned) matrix $A \in \mathbb{R}^{d_{y} \times d_{x}}$ and exogenous noise $\epsilon$. 

However, conventional supervised DL approaches are not ideally suited for practical inverse problems. Large quantities of clean paired data, typically needed for training, are not available in many problem domains, e.g., tomographic reconstruction.
Moreover, ill-posedness (due to the forward operator $A$ and noise $\epsilon$) and high-dimensionality of the reconstructed images $x$ pose significant challenges, and can be computationally very demanding. 
Whereas standard imaging tasks, e.g., denoising and deblurring, use observations of high dimensionality ($d_{x} \approx d_{y}$), tomographic imaging often requires reconstructing an image from observations of a much lower dimensionality.
For example, reconstructing a CT of a walnut (as in \Cref{subsec:walnut}) 
may require reconstructing from observations that are only $3\%$ of the size of the original image.

Unsupervised DL approaches do not require paired training data. Out of these, deep image prior (DIP) \citep{Ulyanov:2018} has garnered the most traction.
DIP parametrises the reconstructed image as the output of a convolutional neural network (CNN) with a fixed input.
The reconstruction process learns low-frequency components before high-frequency ones \citep{chakrabarty2019spectral,ShiSnoek:2022}, which can act as a form of regularisation.

Alas, the practicality of DIP is hindered by two key issues. Firstly, each DIP reconstruction requires training the entire network anew.
This can take from several minutes up to a couple of hours for high-resolution images \citep{barbano2021education}. Secondly, DIP requires careful early stopping \citep{Wang2021stopping} to avoid overfitting, which is often based on case-by-case heuristics.
Unfortunately, validation-based stopping criteria are often not viable in the unsupervised setting due to violated i.i.d. assumptions \cite{Wang2021stopping}.

This paper aims to address both of the existing issues inherent to DIP, and its application to image restoration and reconstruction.
Building upon recent body of evidence showing that neural network (NN) training often takes place in low-dimensional subspaces \citep{Li2018intrinsic, Frankle2019lottery, Tao2022subspace}, we restrict DIP's optimisation to a sparse linear subspace of its parameters. 
This has a two-fold effect. First, subspace optimisation trades off some flexibility in capturing fine image structure details for robustness to overfitting. 
This is extraordinarily well suited in imaging problems belonging to inherently lower dimensional structures, but it is shown to be also competitive in restoring natural images. 
Moreover, it allows using stopping criteria based on the training loss, without sacrificing performance. 
Second, the low dimensionality induced by the subspace allows using second order optimisation methods \citep{Amari2013information, Martens2015kron}.
This greatly reduces running time, and stabilises the reconstruction process, facilitating the use of a simple loss-based stopping criterion.

Our contributions can be summarised as:
\begin{itemize}[topsep=0pt,labelindent=-1pt, parsep=1ex,itemsep=0.5ex]
    \item We extract a principal subspace from DIP's parameter trajectory during a synthetic pre-training stage. To reduce the memory footprint of working with a non-axis-aligned subspace, we sparsify the extracted basis vectors using top-$k$ leverage scoring.
    \item We use second order methods: natural gradient descent (NGD) and limited-memory Broyden– Fletcher–Goldfarb–Shanno (L-BFGS), to optimise DIP's parameters in a low-dimensional subspace. 
    \item We provide thorough experimental results across image restoration and  tomographic tasks of different geometry and degree of ill-posedness, showing that subspace methods are favourable in terms of optimisation stability to reconstruction fidelity trade-off.
\end{itemize}

\section{Deep Image Prior}

DIP expresses the reconstructed image $x = f(x_0, \theta)$ in terms of the parameters $\theta\in\R^{d_\theta}$ of a CNN $f: \R^{d_{x}}\times\R^{\dtheta} \to \R^{d_{x}}$, and fixed input $x_0 \in \R^{d_x}$. 
The parameters $\theta$ are learnt by minimising the loss
\begin{align}\label{eq:dip_tv}
\mathcal{L}(\theta) = \|A f(x_0, \theta) - y \|_2^2+ \lambda {\rm TV}(f(x_0, \theta)),
\end{align}
composed of a data fidelity and total variation (TV), weighed by a constant $\lambda>0$. TV is the most popular regulariser for image reconstruction \citep{rudin1992nonlinear,chambolle2010introduction}. Its anisotropic version is given by
\begin{align}\label{eq:TV_equation}
    \text{TV}(x)\! =\! \sum_{i,j} | X_{i,j} \!-\! X_{i+1,j}| + \sum_{i,j} |X_{i,j} \!-\! X_{i,j+1}|,
\end{align}
where  $X \in \R^{h \times w}$ is a vector $x\in\R^{d_x}$ reshaped into an $h\times w$ image, and $d_{x}{=}h\cdot w$. In this work, $f$ is a fully convolutional U-Net, see \ref{app:arch} for more details. 
When optimising $f$ from random initialisation, \citep{chakrabarty2019spectral} and \citep{ShiSnoek:2022} find that low-frequency image components are learnt faster than high-frequency ones.
This implicitly regularises the reconstruction by preventing overfitting to noise, as long as the optimisation is stopped early enough.

DIP optimisation costs can be often reduced by pre-training on synthetic data. 
E-DIP \citep{barbano2021education} first generates samples from a traning data distribution $P$ of random ellipses, and then applies the forward model $A$ and adds white noise, following \cref{eq:inverse_problem}.
The network input is set as the filtered back-projection (FBP)
$x_0\!=\!x^\dagger\! \coloneqq\! A^{\dagger}y$, where $A^{\dagger}$ denotes the (approximate) pseudo-inverse of $A$. 
The pre-training loss $\mathcal{L}_{\rm pre}(\theta)$ is given by
\begin{equation}\label{eq:pre_training_objective}
    \mathcal{L}_{\mathrm{pre}}(\theta) = \mathbb{E}_{x,y\sim P} \|f(x^\dagger, \theta) -  x  \|_2^2,
\end{equation}
The pre-trained network can then be fine-tuned on any new observation $y$ by optimising the objective \cref{eq:dip_tv} with FBP as the input.
E-DIP decreases the DIP's training time, but can increase susceptibility to overfitting, making early stopping even more critical, cf. discussion in \cref{sec:experiments}.

\section{Methodology}\label{sec:method}

We now describe our procedure for DIP optimisation in a subspace. We first describe how the E-DIP pre-training trajectory is used to extract a sparse basis for a low-dimensional subspace of the parameters. 
The objective is then reparametrised in terms of sparse basis coefficients.
Finally, we describe how L-BFGS and NGD are used to update the sparsified subspace coefficients.

\paragraph{Step 1: Identifying the sparse subspace}\label{subsec:identifying_subspace}

We find a subspace of parameters that is low-dimensional and \textit{easy to work with}, but contains a rich enough set of parameters to fit to the observation $y$. 
We leverage E-DIP pre-training trajectory to acquire basis vectors by stacking $\dpre$ parameter vectors, sampled at uniformly spaced checkpoints on the pre-training trajectory, into a matrix $\Theta^{\mathrm{pre}} \in \R^{\dtheta \times \dpre}$. 
We then compute top-$\dsub$ SVD of $\Theta^{\mathrm{pre}} \approx U S V^\top$, and keep the left singular vectors $U \in \R^{\dtheta \times \dsub}$, where $\dsub \le \dpre$ is the dimensionality of the chosen subspace. We then sparsify the orthonormal basis $U$ by computing leverage scores \citep{Drineas2012leverage}, associated with each DIP parameter as
\begin{equation*}
    \ell_{i} = \sum_{k=1}^{\dsub} [U]_{ik}^{2},\quad i=1,\ldots, d_\theta.
\end{equation*}
We keep the basis vector entries corresponding to $\dlev < \dtheta$ largest leverage scores. This can be achieved by applying a mask $M \in \{0, 1\}^{\dtheta \times \dtheta}$ satisfying $[M]_{ii} = \mathbbm{1}(i \in \textrm{arg top-}\dlev  \: \ell)$, where $\ell = [\ell_1,\ell_2,\dotsc,\ell_{d_{\theta}}]$. The sparse basis $M U$ contains at most $\dlev \cdot\dsub$ non-zero entries.

Pre-training and sparse subspace selection are only performed once, and can be amortised across different reconstructions. We choose $d_{\mathrm{pre}}\!=\!\mathcal{O}(10^3)$, resulting in a large memory footprint of the matrix $\Theta^{\mathrm{pre}}$, though this is stored in cpu memory. Alternatively, ncremental SVD algorithms \cite{brand2002incremental} can be used to further reduce the memory requirements (see \cref{app:additional_results}).
Training DIP in the sparse subspace requires storing only the sparse basis vectors $MU$ in accelerator (gpu / tpu) memory.
Thus, sparsification allows training in relatively large subspaces $\dsub\,{>}\,10^3$ of large networks $\dtheta\,{>}\,10^7$.

\paragraph{Step 2: Network reparametrisation}\label{subsec:reparametrisation}

We reparametrise the objective $\mathcal{L}(\theta)$ in terms of sparse basis coefficients $c \in \R^{\dsub}$ as
\begin{align}\label{eq:objective_reparametrisation}
    \mathcal{L}_{\gamma}(c) \coloneqq \mathcal{L}(\gamma(c)),\quad \text{with }  \gamma(c) \coloneqq \theta^{\mathrm{pre}} + M U\, c.
\end{align}
This restricts the DIP parameters $\theta^{\mathrm{pre}}$ (from pre-training) to change only along the sparse subspace.
The coefficient vector $c$ is initialised as a sample from a uniform distribution on the unit sphere. 
\paragraph{Step 3: Second order optimisation}\label{subsec:2O_optimisation}
The reparametrised subspace DIP objective $\mathcal{L}_{\gamma}$ in \cref{eq:objective_reparametrisation} differs from traditional deep learning loss functions in that
(i) deterministic estimates of $\mathcal{L}_{\gamma}$ and its gradients can be computed from a single pass through the network; (ii) the local curvature matrix of $\mathcal{L}_{\gamma}$ can be computed and stored without resorting to approximations.
These facts open the door to second order optimisation of $\mathcal{L}_{\gamma}$, which may converge faster than first order methods. However, the cost of repeatedly evaluating second order derivatives of neural networks is prohibitive and this is compounded by the rapid change of the local curvature of the non-quadratic loss when traversing the parameter space.
We mitigate this by performing online low-rank updates of a curvature estimate while only accessing loss Jacobians. In particular, we use L-BFGS \citep{lbfgs1989} and stochastic NGD \citep{amari1998natural,Martens2022review}. The former estimates second directional derivatives by solving the secant equation. The latter keeps an exponentially moving average of stochastic estimates of the Fisher information matrix (FIM).
\paragraph{NGD for DIP in a subspace}

The exact FIM is given as
\begin{gather}\label{eqn:Fisher}
  \mathbb{E}_{v \sim\mathcal{N}(Af(x^{\dagger}, \gamma(c)), \, I_{d_y})}[\nabla_{c} \mathcal{L}_{\gamma}(c)^\top \nabla_{c} \mathcal{L}_{\gamma}(c)], %
\end{gather}
where $\nabla_{c} \mathcal{L}_{\gamma}(c) = (A f(x^\dagger, \gamma(c)) - v)^{\top} AJ_{f}MU\in \R^{1\times\dsub}$ is the Jacobian of the subspace loss $\mathcal{L}_\gamma$ at the current coefficients, and $J_f \coloneqq \nabla_{\theta} f(x^{\dagger}, \theta)|_{\theta = \gamma(c)} \in \R^{d_{x} \times \dtheta }$ is the network Jacobian at $\theta=\gamma(c)$. 
At step $t$ %
we estimate the FIM by Monte-Carlo
\begin{gather}\label{eqn:fisher_probing}
\hat{F}_{t} = \frac{1}{n} \sum_{i=1}^n (z_i^\top A J_f M U)^\top z_i^\top A J_f M U ,\quad \text{with } z_{i} \sim \mathcal{N}(0, I_{d_{y}}).
\end{gather}
The moving average FIM is then updated as
\begin{align}\label{eq:FMA}
    F_{t+1} &= \beta F_t + (1 - \beta) \hat{F}_{t} \quad \textrm{with } \beta \in (0,1).
\end{align}
TV's contribution is omitted since the FIM is defined only for terms that depend on the observations and not for the regulariser \citep{ly2017tutorial}.
Our implementation of NGD follows the approach of
\citep{Martens2015kron}, with adaptive damping, and step size and momentum parameters chosen by locally minimising a quadratic model. See \cref{app:NGD} for additional discussion.

\begin{figure*}[t]
\vspace{-0.05cm}
    \centering
    \includegraphics[width=0.96\linewidth]{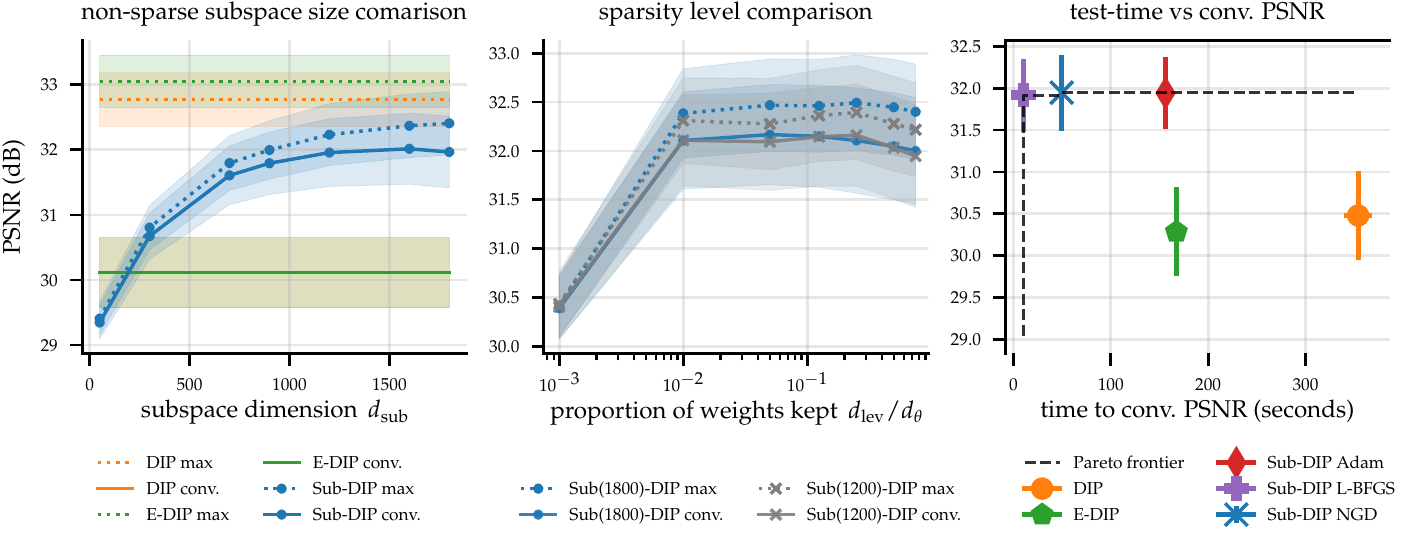}
    \caption{
    The influence of subspace dimension $d_{\rm sub}$ (left), sparsity level $\nicefrac{d_{\rm lev}}{d_\theta}$ (middle) on PSNR, and the PSNR vs time Pareto-curve (right) for CartoonSet. ``max'' refers to the oracle stopping PSNR, while ``conv'' refers to the PSNR at the stopping point found by applying \cref{alg:1}, cf. \cref{app:additional_results}, to the loss \eqref{eq:dip_tv}, with $\delta{=}0.995$ and patience of $\mathfrak{p}{=}100$. Left and middle plots use NGD. Results show mean and standard deviation over 25 reconstructed images. Runs are performed on A100 GPU.
    }
    \vspace{-0.15cm}
    \label{fig:cartoon_ablation}
\end{figure*}

\section{Experiments}\label{sec:experiments}

Our experiments cover a wide range of image restoration and tomographic reconstruction tasks, two challenging classes of imaging problems. 
In \cref{subsec:cartoon}, we conduct an ablation study on CartoonSet \cite{royer2017xgan}, examining the impact of subspace dimension $\dsub$, subspace sparsity level $\dlev$, problem ill-posedness, and choice of the optimiser on reconstruction speed and fidelity.
The acquired insights are then applied to real-life tomography tasks and image restoration. We compare the reconstruction fidelity, propensity to overfitting, and convergence stability relative to vanilla DIP and E-DIP on a real-measured high-resolution CT scan of a walnut, in \cref{subsec:walnut}, and on medically realistic high-resolution abdomen scans from the Mayo Clinic, in \cref{subsec:mayo}. 
We simulate observations using \cref{eq:inverse_problem} with dynamically scaled Gaussian noise given by
\begin{equation}\label{eqn:dynamic_noise}\epsilon \sim \mathcal{N}(0, \,\sigma^2 I_{d_{y}}) \quad \text{with }\sigma = p/d_{y}\sum_{i=1}^{d_y} |y_i|,
\end{equation}
with the noise scaling parameter set to $p=0.05$, unless noted otherwise.
We conclude with denosing and deblurring on a standard RGB natural image dataset (Set5) in  \cref{sec:natural_images}.

For studies in Sections \ref{subsec:walnut}, \ref{subsec:mayo} and \ref{sec:natural_images}, we use a standard fully convolutional U-Net architecture with either $\sim 3$M (for CT tasks) or $\sim 1.9$M (for natural images) parameters, see architectures in \cref{app:arch}.
For the ablative analysis in \cref{subsec:cartoon}, we use a shallower architecture ($\sim .5$M) with only 64 channels and four scales, keeping the skip connections in lower layers. 

Following the literature, we use Adam to train the vanilla DIP \citep{Ulyanov:2018} (labelled DIP) and E-DIP \citep{barbano2021education}.
We train subspace coefficients with Adam (Sub-DIP Adam) as a baseline, L-BFGS (Sub-DIP L-BFGS) and NGD (Sub-DIP NGD). 
Image quality is assessed through peak signal-to-noise ratio (PSNR).

For tomographic tasks we use the same pre-training runs for E-DIP and all subspace methods: minimising \cref{eq:pre_training_objective} over 32k images of ellipses with random shape, location and intensity. Pre-training inputs are constructed from an FBP obtained with the same tomographic projection geometry as the dataset under consideration. 
Analogously, for image restoration tasks, we pretrain on ImageNet \cite{deng2009imagenet}. 

The method is built on top of the E-DIP library ({\href{https://github.com/educating-dip/educated_deep_image_prior}{\texttt{github.com/educating-dip}}}). The full implementation and data are available at \href{https://github.com/anonsubdip/subspace_dip}{github.com/anonsubdip/subspace\_dip}.

\subsection{Ablative analysis on CartoonSet \cite{royer2017xgan}} \label{subsec:cartoon}

\begin{figure}
\includegraphics[width=\linewidth]{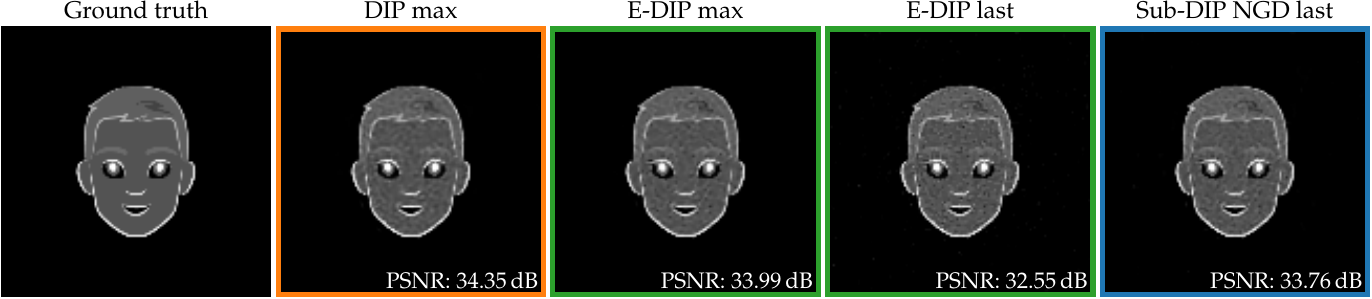}
\caption{Reconstruction comparison for an example CartoonSet image from 45 angles. ``max'' indicates oracle (highest possible) PSNR. ``last'' denotes the final reconstruction.
}
\label{fig:cartoon_example_recon}
\end{figure}

We investigate Sub-DIP's sensitivity to subspace dimension $\dsub$, sparsity level $\dlev$, and  ill-posedness on 25 images of size $(128\,\text{px})^2$ from CartoonSet \cite{royer2017xgan} ({\href{https://google.github.io/cartoonset/}{\texttt{google.github.io/cartoonset}}}). 
Example reconstructions are shown in Fig. \ref{fig:cartoon_example_recon}.
We simulate a parallel-beam geometry with 183 detectors and $45$, $95$ or $285$ angles, corresponding to, respectively, a very sparse-view ($d_y{=}8235$), a moderately sparse-view ($d_y{=}17385$), and a fully sampled setting ($d_y{=}52155$). We construct subspaces by sampling $\dpre{=}2k$ parameters at uniformly spaced checkpoints during 100 pre-training epochs on  ellipses. We measure the degree of overfitting by comparing the highest PSNR obtained throughout optimisation (max) with that given by \cref{alg:1}, applied to the training loss \eqref{eq:dip_tv} with $\delta{=}0.995$ and $\mathfrak{p}{=}100$ steps (conv.). Further ablation on SVD extraction and quality of used basis are reported in \cref{app:additional_results}.

\paragraph{Subspace dimension} 

Fig. \ref{fig:cartoon_ablation} (left) explores the trade-off between subspace dimension $\dsub$ and reconstruction quality.
We use 95 angles, subspace methods with no sparsity ($\dlev{=}d_\theta$) and the NGD optimiser.
Both standard DIP and E-DIP overfit, showing a $\approx 3$ dB gap between max and conv. PSNR values, while subspace methods exhibit only a $\approx 0.5$ dB gap.
Both max and conv. PSNR present a monotonically increasing trend with subspace dimension, while the gap at $d_{\rm sub } > 1k$ stays roughly constant at $\sim$0.25 dB. Thus, these subspace dimensions are too small for significant overfitting to occur.
In spite of this, $\dsub{=}100$ is enough to obtain better conv. PSNR than DIP and E-DIP. 

\paragraph{Subspace sparsity}
Fig. \ref{fig:cartoon_ablation} (middle) shows that for $\nicefrac{\dlev}{d_\theta}>0.01$, the reconstruction fidelity is largely insensitive to the sparsity level. 
This is true for both $d_{\rm{sub}}{=}1200$ and $d_{\rm{sub}}{=}1800$. This effect is also somewhat independent of $d_{\rm sub}$.
Hence, \emph{sparse subspaces should be constructed by choosing a large $d_{\rm sub}$ and then sparsifying its basis vectors to ensure computational tractability.}

\begin{wrapfigure}{r}{0.5\textwidth}
\begin{center}
\vspace{-0.7cm}
    \includegraphics[width=\linewidth]{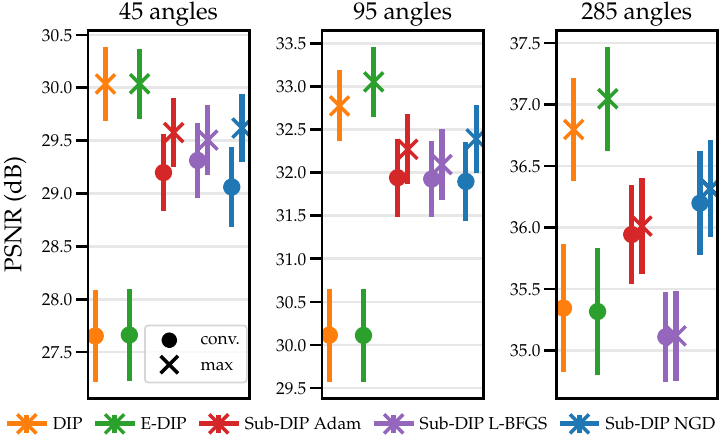}
    \vspace{-0.55cm}
    \caption{PSNR mean and standard deviation over 50 CartoonSet images from 45, 95, and 285 angles.}
    \label{fig:cartoon_overfitting}
\end{center}
\end{wrapfigure}

\paragraph{Problem ill-posedness}
We report the performance with varying numbers of view angles in Fig. \ref{fig:cartoon_overfitting}, with non-sparse subspaces of dimension $d_{\rm{sub}}{=}1800$. 
Subspace methods present a smaller gap between max and conv. PSNR ($\sim0.25$ dB) than DIP and E-DIP ($\sim 2.5$ dB). 
Subspace methods also present better fidelity at convergence for all the studied number of angles.
Notably, the improvement is larger for more ill-posed settings ($ \sim 2.5$ dB at 45 and 95 vs $\approx 1$ dB at 285 angles), despite the worsening performances for all methods in sparser settings. 
This is expected as there is a reduced risk of overfitting in data-rich regimes and more flexible models can do better. 
E-DIP's max reconstruction fidelity is consistently above that of other methods by at least $0.5$ dB.
This may be attributed to full-parameter flexibility with benign inductive biases from pre-training. However, obtaining max PSNR performance requires oracle stopping and is not achievable in practice.

\paragraph{First vs second order optimisation}

We compare optimisers' conv. PSNR vs their time to convergence. 
Fig. \ref{fig:cartoon_ablation} (right) shows that Sub-DIP L-BFGS and NGD converge in less than $50$ seconds. 
These methods are Pareto-optimal, with the former reaching $\sim\!2\!$ dB higher reconstruction fidelity and the latter converging faster (in $\sim\!20$ seconds). Sub-DIP Adam retains protection against overfitting but converges at a rate similar to non-subspace first order methods (in $\sim 180$ seconds). 
These trends hold across studied degrees of ill-posedness; see  \cref{app:additional_cartoon}.

\begin{figure}[H]
\vspace{-0.0cm}
    \centering
    \includegraphics[width=\linewidth]{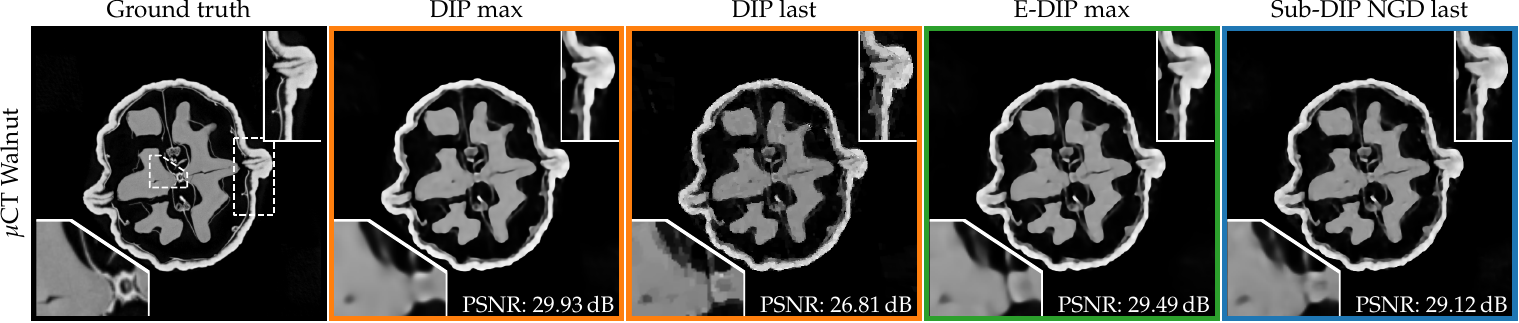}
    \vspace{-0.55cm}
    \caption{Reconstructions of the Walnut using 60 angles. Sub-DIP reconstructions do not capture any noise, but present slightly increased ringing around very thin structures.}
    \label{fig:walnut_images}
    \vspace{-0.1cm}
\end{figure}

\subsection{\textmu CT Walnut dataset \cite{der_sarkissian2019walnuts_data}}\label{subsec:walnut}

Now we study the methods on a real-measured high-resolution \textmu CT problem.
We reconstruct a $(501\,\text{px})^2$ slice from a very sparse, real-measured cone-beam scan of a walnut, using 60 angles and 128 detector pixels ($d_{y}=7680$).
We compare reconstructions against ground-truth \citep{der_sarkissian2019walnuts_data}, which uses classical reconstruction from full data acquired at 3 source positions with 1200 angles and 768 detectors.
This task involves fitting both broad-stroke and fine-grained image details (see Fig. \ref{fig:walnut_images}), making it a good proxy for \textmu CT of industrial context.
We pre-train the 3M parameter U-Net for 20 epochs and save $\dpre{=}5k$ parameter checkpoints. Following \cref{subsec:cartoon}, we construct a $\dsub{=}4k$ dimensional subspace and sparsify it down to $\nicefrac{\dlev}{d_{\theta} }{=}0.5$ of the parameters.

Fig. \ref{fig:tomography_convergence} (left) shows the optimisation curves for all optimisers, averaged across 3 seeds. Qualitatively, the results are similar to the cartoon data. 
Second order subspace methods converge to their highest reconstruction fidelity within the first $500$ seconds and do not overfit. 
Vanilla DIP and E-DIP also converge quickly but suffer from overfitting leading to ${\sim}3$ dB and $\sim 1.8$ dB of performance degradation, respectively.
Sub-DIP Adam takes over $3000$ seconds to converge and does not overfit.

\paragraph{Stopping criterion challenges}

\begin{wrapfigure}{r}{0.5\textwidth}
\begin{center}
\vspace{-0.75cm}
\includegraphics[width=\linewidth]{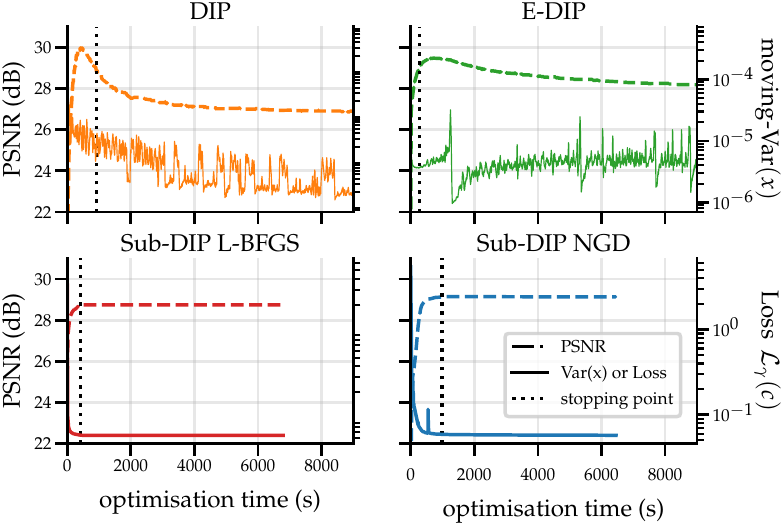}
\vspace{-0.4cm}
\caption{
The evolution of PSNR and stopping metrics (variance-based for DIP and E-DIP and loss-based for Sub-DIP NGD and L-BFGS.) vs optimisation time on Walnut with one seed.
}
\end{center}
\label{fig:walnut_stopping}
\end{wrapfigure}
To capture oracle performance of DIP and E-DIP, one would require a robust stopping criterion.
Note that we cannot base our stopping criterion on \cref{eq:dip_tv}.
We instead turn to the method from \citep{Wang2021stopping} which minimises a rolling estimate of the reconstruction variance across optimisation steps, a proxy for the squared error. Following \citep{Wang2021stopping}, we compute variances with a 100 step window and apply \cref{alg:1} with a patience of $\mathfrak{p}{=}1000$ steps and $\delta{=}1$.
\Cref{fig:walnut_stopping} shows this metric to be very noisy when applied to DIP and E-DIP.
This breaks the smoothness assumption implicit in the stopping criterion (\cref{alg:1}), leading to stopping more than $1$ dB before/after reaching the max PSNR.
This is due to the severe ill-posedness of the tasks causing variance across a large subspace of reconstructions that fit our observations well.
For tomographic reconstruction, the variance curve becomes more non-convex, and its minimum tends to present a shift relative to the optima of the reconstruction PSNR.
Since subspace methods don't overfit and the loss is smooth, we can use it as our stopping metric ($\delta{=}0.995$, $\mathfrak{p}=100$).

\begin{figure*}[t]
\vspace{-0.15cm}
\begin{center}
    \includegraphics[width=0.95\linewidth]{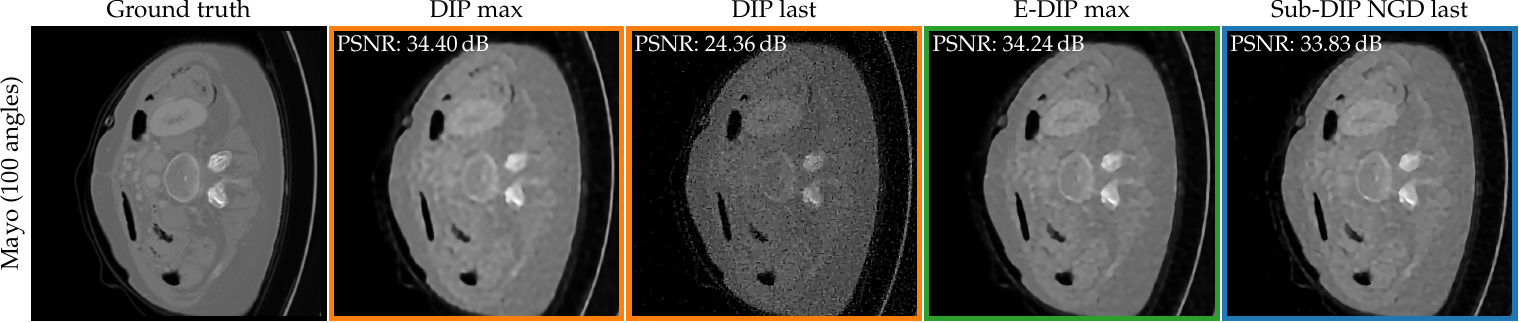}
    \\
    \includegraphics[trim={0 0 0 0.31cm},clip, width=0.95\linewidth]{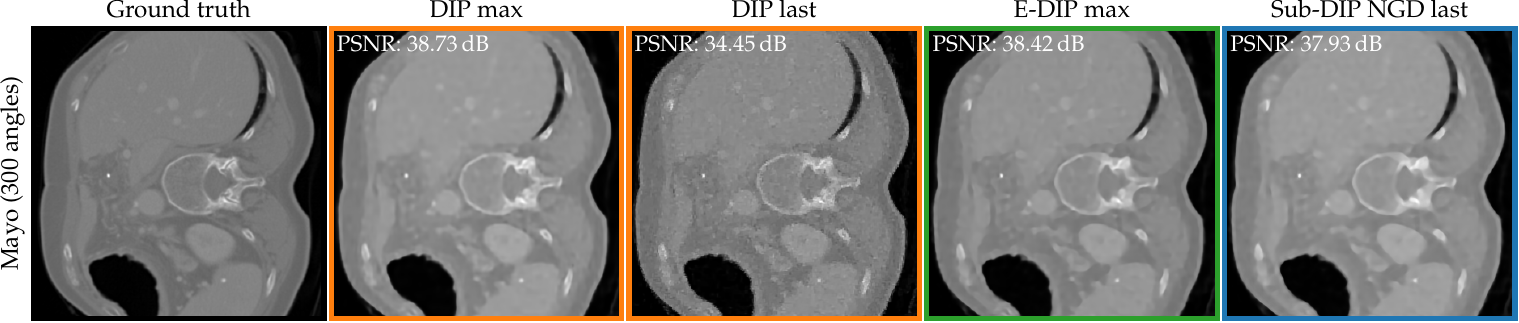}
\end{center}
    \caption{
Example reconstructions on the Mayo dataset in 100 (top) and 300 (bottom) angle settings.}
    \label{fig:mayo_reconstruction_maintext}
\end{figure*}

\subsection{Mayo Clinic dataset  \cite{moen2021mayo_ldct_and_projection} }\label{subsec:mayo}
\begin{figure*}[ht]
    \centering
    \includegraphics[width=0.96\linewidth]{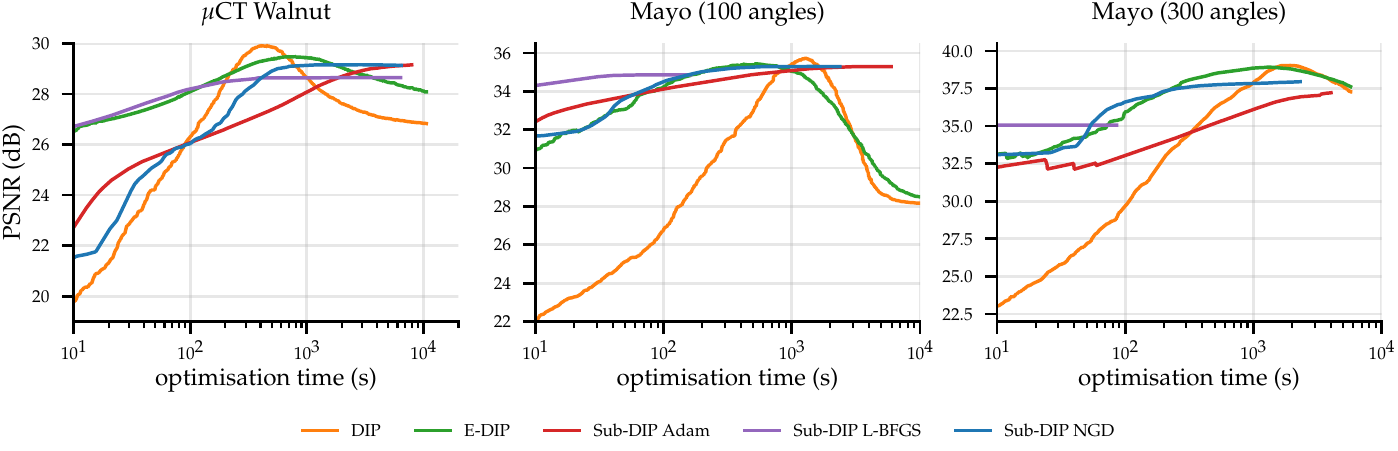}
    \vspace{-0.2cm}
    \caption{
    Optimisation curves for the $\mu$CT Walnut data (left), and Mayo using 100 angles (middle) and Mayo using 300 angles (right) averaged over 10 images.
    }
    \label{fig:tomography_convergence}
\end{figure*}
To investigate a medical setting, we use 10 clinical CT images of the human abdomen released by Mayo Clinic \citep{moen2021mayo_ldct_and_projection} and study the behaviour of subspace optimisation in a medical image class.
We reconstruct from a simulated fan-beam projection, with 300 angles and white noise with the noise scaling parameter $p=0.025$, cf. \eqref{eqn:dynamic_noise}.
As a more ill-posed reference setting, we use 100 angles (comparable sparsity to the Walnut setting) and Gaussian noise with $p{=}0.05$.
For both tasks, we use the $3$M parameter U-Net, as in \cref{subsec:walnut}, pre-trained on 32k ellipse samples. 
For the sparse setting (100 angles), we use a $\dsub=4$k dimensional subspace constructed from $\dpre{=}5$k checkpoints, but $\nicefrac{\dlev}{d_{\theta}}{=}0.25$. 
For the more data-rich setting (300 angles), we use $\dsub=8$k, sampled from $\dpre{=}10$k checkpoints, and similarly, we sparsify it down to $\nicefrac{\dlev}{d_{\theta}}{=}0.25$.

In the of 300 angle setting, Sub-DIP NGD reaches $37$ dB within 200 seconds and then further increases the PSNR but slowly, without performance saturation, cf. Fig. \ref{fig:tomography_convergence} (right).
In contrast, for the sparser Walnut and Mayo data, Sub-DIP NGD maintains a steep PSNR increase until reaching max PSNR. Interestingly, L-BFGS does not perform well in the 300 angle setting, obtaining $< 36$ dB PSNR. 
This might be due to L-BFGS's tendency to stop the iterations too early in high dimensions.

Fig. \ref{fig:mayo_reconstruction_maintext} shows example reconstructions using 100 and 300 angles. If stopped within a narrow max PSNR window, DIP and E-DIP can deliver reconstructions that better capture high-frequency details than Sub-DIP methods as one would expect. However, while the DIP and E-DIP reconstructions become noisy once they start to overfit, Sub-DIP methods do not exhibit any noise.
We deem the \textit{increased robustness vs reduced flexibility} trade-off provided by the Sub-DIP to be favourable, even in the well-determined setting.

\begin{figure}[ht]
	\begin{minipage}[c][1\width]{
	   0.48\textwidth}
    \vspace{-0.0cm}
    \centering
    \includegraphics[keepaspectratio]{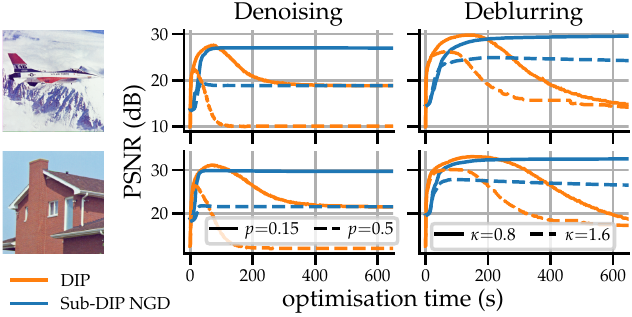}
    \vspace{-0.25cm}
\end{minipage}\hspace{0.05em}
	\begin{minipage}[c][1\width]{
	   0.48\textwidth}
    \vspace{-0.0cm}
    \centering
    \includegraphics[keepaspectratio]{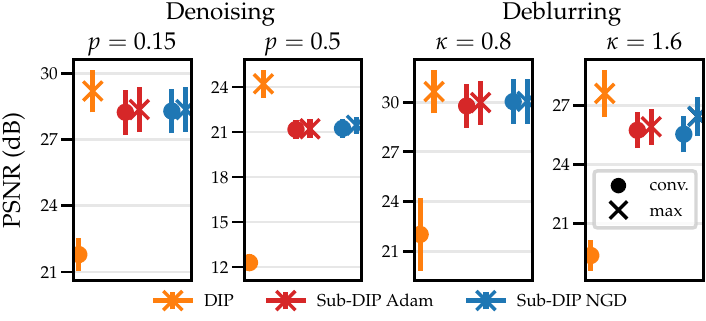}
    \vspace{-0.675cm}
\end{minipage}
    \vspace{-1.5cm}
\caption{Denoising ($p=\{0.15,0.5\}$) and deblurring ($\kappa=\{0.15,0.5\}$) on the ``jet F16'' and ``house'' images. On the left we show PSNR trajectories for each of the tasks, comparing DIP and Sub-DIP NGD, and on the right we show mean and std of max and conv. PSNR over the studied 5 images. 
Results for the other two noise levels are reported in \cref{app:additional_results}.
}
\label{fig:natural_images}
\end{figure}

\subsection{Image restoration of Set5}\label{sec:natural_images}

We conduct denoising and deblurring on five widely used RGB natural images (``baboon'', ``house'', ``jet F16'', ``Lena'', ``peppers'') of size $(256\,\text{px})^2$. 
The pre-training is done on ImageNet \cite{deng2009imagenet}, a dataset of natural images, which we use to extract the basis for the subspace methods.

For denoising we consider four noise settings with the noise scaling parameter, cf. \eqref{eqn:dynamic_noise}, i.e. $p\in\{0.10, 0.15, 0.25, 0.5\}$. 
We extract a single subspace for all the noise levels.
To this end, during the pre-training stage, we construct a dataset by adding noise to each training datum, with a randomly selected noise scaling parameter $p\sim\rm{Uni}(0.05,0.5)$.
Then a $\dsub=8k$ subspace with sparsity level $\dlev/\dtheta=0.5$ is extracted from $\dpre=10k$ samples.
In the high noise case when $p=0.5$, we further sub-extract to a smaller $\dsub=1k$ subspace.
For deblurring we consider two settings, using a Gaussian kernel with std of $\kappa\in\{0.8, 1.6\}$ pixels and $p=0.05$ Gaussian noise. We then follow an analogous procedure; adding $p=0.05$ Gaussian noise and applying $\kappa\sim\rm{Uni}(0.4,2)$ blur to each training datum, and then constructing a single subspace.

As common practice when deploying the DIP on restoration tasks, we do not include the TV ($\lambda = 0$) in \eqref{eq:dip_tv} for neither denoising nor deblurring. Instead, \textit{the regularising property of the reconstruction stems exclusively from restricting the DIP optimisation to a low-dimensional subspace of its parameters}.
\begin{figure}
    \centering 
    \includegraphics[width=\linewidth]{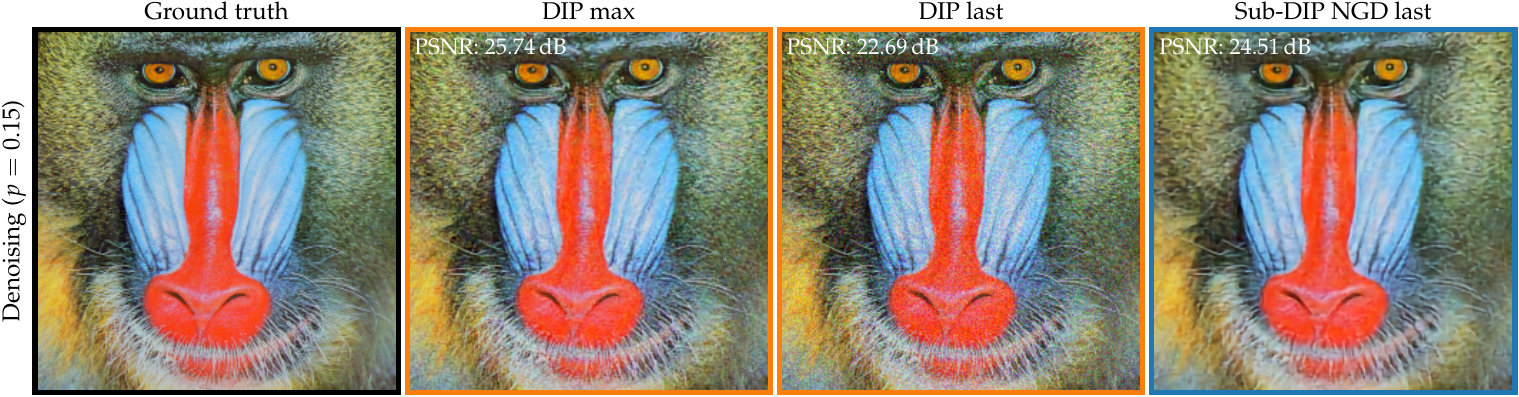}
    \caption{Restoration of the noisy (with noise scaling parameter $p=0.15$) ``baboon'' image.}
    \label{fig:baboon}
\end{figure}

Fig. \ref{fig:natural_images} (left) shows PSNR trajectories for denoising and deblurring on ``jet F16'' and ``house'' images.
These images contain large regions of continuous colour intensity with clearly defined borders between regions. Hence, we expect subspace methods to perform well. This is confirmed by the results: Sub-DIP NGD and DIP have a comparable max PSNR.
In Fig. \ref{fig:natural_images} (right) we show the mean and
std of max and conv. PSNR over 5 studied natural images for low and high noise and blur.

Fig. \ref{fig:baboon} shows the reconstruction for denoising on the ``baboon'' image with a moderate noise level ($p=0.15$).
The ``baboon'' image contains fine-grained details and sharp transitions throughout the image. 
Hence, we expect a somewhat comparatively worse performance of subspace methods on this task, as higher frequency information can be lost when using low-dimensional approaches.
The results confirm this intuition: the max PSNR for the ``baboon'' with E-DIP is at least $1$ dB higher than that for subspace methods.
Moreover, while standard DIP experiences a sharp decline after achieving max PSNR, subspace methods retain the performance.

\section{Related Work}

The present work lies at the intersection of unsupervised methods, subspace learning and second order optimisation. We discuss related work in each of these fields.

\paragraph{Avoiding overfitting in DIP}
Since the introduction of the DIP \citep{Ulyanov:2018,ulyanov2020dip}, stopping optimisation before overfitting noise has been a necessity. The analysis in  \citep{chakrabarty2019spectral} and \citep{ShiSnoek:2022} elucidate that U-Net is biased towards learning low-frequency components before high-frequency ones. The authors suggest a sharpness-based stopping criterion, which however requires a modified architecture.
\citep{Yeonsik2021denoising} propose a criterion based on the Stein’s unbiased risk estimate \citep{eldar2008generalized} for denoising, but which performs poorly for ill-posed settings \citep{metzler2018unsupervised}.
\citep{Wang2021stopping} propose a running image-variance estimate as a proxy for the reconstruction error. Our experiments find this method somewhat unreliable for sparse CT reconstruction.
\citep{ding2022validation} and 
\citep{yaman2021zeroshot} propose to split the observation vector into training and validation sub-vectors, and use loss on the latter as a stopping criteria.
Unfortunately, this violates the i.i.d. data assumption that underpins validation-based early stopping (Theorem 11.2, \citep{learningtheorybook}).
Independently, 
\citep{LiuSunXuKamilov:2019} and \citep{baguer2020diptv} add a TV regulariser to the DIP objective \cref{eq:dip_tv}. This only partially alleviates the need for early stopping and has seen widespread adoption. To the best of our knowledge, the present work is the first to successfully avoid overfitting without significant performance degradation.

\paragraph{Linear subspace estimation} Literature on low-rank matrix approximations is rich, with randomised SVD approaches being the most common \citep{Halko2011structure, MartinssonTropp:2020}. However, in high-dimensions, even working with a small set of dense basis vectors can itself be prohibitively expensive.
Matrix sketching methods \citep{Drineas2012leverage,Liberty2013sketching} alleviate this through axis-aligned subspaces. To the best of our knowledge, our work is the first to combine these two method classes, producing non-axis-aligned but sparse approximations.

\paragraph{Optimising neural networks in subspaces}

Closely related to the present work is that of \citep{Li2018intrinsic} and \citep{Wortsman2021learning}, who find that networks can be trained in low-dimensional subspaces of the original parameters without loss of performance, and that more complex tasks need larger subspaces. Similarly to our methodology, Tao et al \citep{Tao2022subspace} identify subspaces from training trajectories and note that this results in robustness to label noise.
\citep{Frankle2019lottery} presents similar findings when large numbers of parameters are ablated, i.e., learning can occur in axis-aligned subspaces. This principle has yielded speedups in network evaluation \citep{Wen16Structured, Daxberger21subnetwork}. 
\citep{Shwartz2022pretrain} use a low-rank estimate of the curvature around an optimum of a pre-training task to regularise subsequent supervised~learning.

\paragraph{Second order optimisation for neural networks}

Despite their adoption in traditional optimisation \citep{lbfgs1989}, second order methods are rarely used with neural networks due to the high cost of dealing with curvature matrices for high-dimensional functions.
\citep{Martens2012free} use truncated conjugate-gradient to approximately solve against a network's Hessian. However, a limitation of the Hessian is that it is not guaranteed to be positive semi-definite (PSD). This is one motivation for NGD \citep{Foresee1997Bayes,Amari2013information,Martens2022review}, that uses the FIM (guaranteed PSD). Commonly, the KFAC approximation \citep{Martens2015kron} is used to reduce the costs of FIM storage and inversion.
Also, common deep-learning optimisers, e.g., Adam \citep{kingma2014method} or RMSprop \citep{rmsprop} may be interpreted as computing online diagonal approximations to the Hessian.

\section{Conclusion}

In this work, we develop a novel approach that constrains the DIP optimisation to a principal low-dimensional subspace, extracted from pre-training trajectories. This greatly reduces, if not completely eliminates, overfitting.
Our approach may be understood from the perspective of the bias-variance tradeoff. At initialisation, the vanilla DIP presents a useful bias towards learning low-frequency image components; but its over-parameterisation leads to overfitting. Pre-training only partially removes DIP’s low pass bias, allowing the E-DIP to often fit images quickly. 
This comes at the cost of increased variance. Our approach seeks to efficiently navigate the bias vs. variance tradeoff. Constraining the optimisation to a low-dimensional subspace greatly reduces variance. By extracting the subspace from the principal directions of pre-training trajectories, and through the use of leverage scoring, we limit the bias introduced into our model.
Furthermore, optimising in lower dimensional subspaces allows using fast and stable second order optimisers. 

In our experiments experiments on several image restoration and tomographic tasks, subspace DIP methods deliver reconstructions on par with DIP’s max performance, and result in better reconstruction quality than the overfit DIP reconstructions.
We aim for this approach to bring DIP-based CT reconstruction methods closer to co-located deployment in real-life settings.

\bibliography{NeurIPS/bibliography}

\newpage
\appendix
\onecolumn

\section{Additional experimental results}\label{app:additional_results}

In this section, we complement the experimental investigation of \cref{sec:experiments} with additional results.

\subsection{Additional CartoonSet results and discussion}\label{app:additional_cartoon}

\paragraph{Average Optimisation Trajectories}

\begin{figure*}[htb]
\vspace{-0.05cm}
    \centering
    \includegraphics[width=0.65\linewidth,trim=0cm 0cm 0cm 0cm,clip]{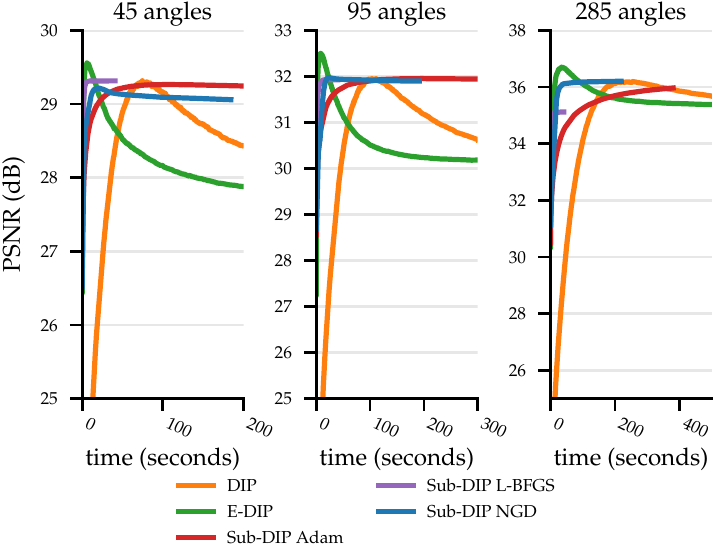}
    \caption{Average PSNR trajectories for 45, 95, and 285 angles over 50 reconstructions.
    }
    \vspace{-0.15cm}
    \label{fig:cartoon traces}
\end{figure*}

In \cref{fig:cartoon traces}, we show PSNR trajectories of the studied methods (DIP, E-DIP, Sub-DIP Adam,  Sub-DIP LBFGS, and Sub-DIP NGD) for 45, 95, and 285 angles, throughout the optimisation. The results, which are averaged over 50 reconstructed images, offer additional evidence supporting previous observations. While DIP and E-DIP tend to overfit to noise once they reach their maximum PSNR values, subspace methods consistently maintain stable reconstruction performance without any noticeable degradation.

\paragraph{Pareto frontier}

\begin{figure*}[htb]
\vspace{-0.05cm}
    \centering
    \includegraphics[width=\linewidth,trim=0cm 0cm 0cm 0cm,clip]{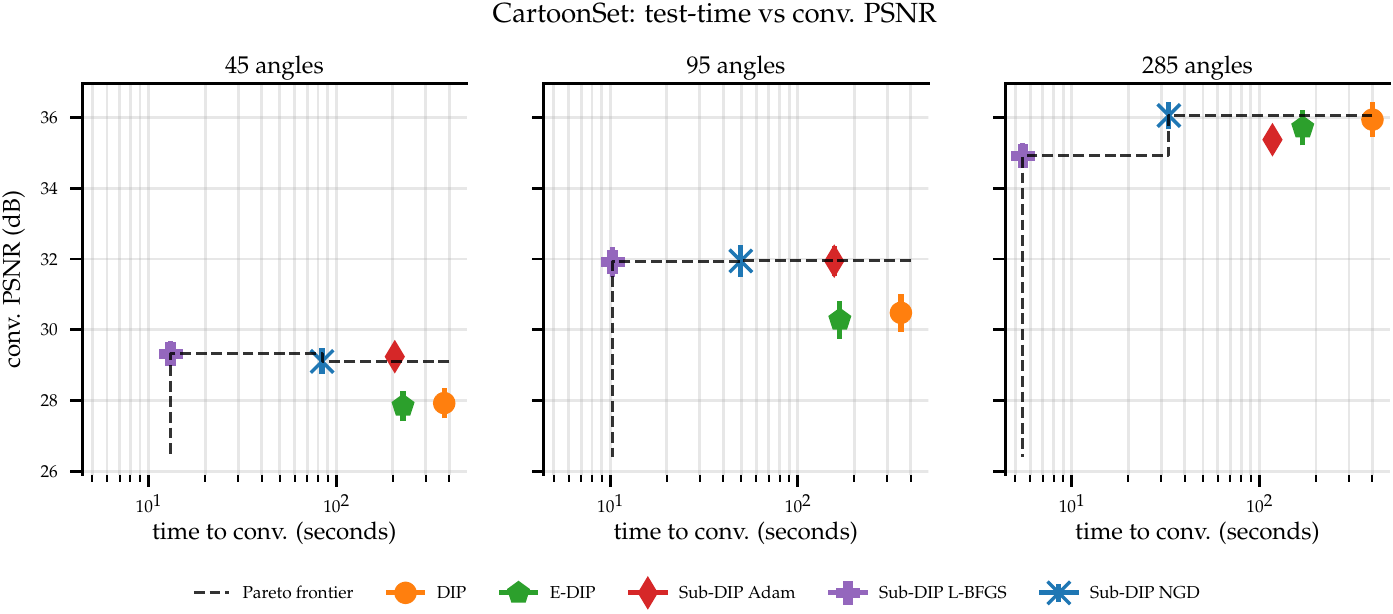}
    \caption{Pareto-curves of conv. PSNR vs optimisation time on the CartoonSet for reconstructions from 45 angles (left), 95 angles (middle) and 285 angles (right). We provide mean and standard deviation of the PSNR, computed across 50 cartoon images. Note that the x-axis is given in log-scale and that the std is hard to observe due to the shared range along the y-axis.
    }
    \vspace{-0.15cm}
    \label{fig:appendix_cartoon_pareto}
\end{figure*}

\begin{figure*}[htb]
\vspace{-0.05cm}
    \centering
    \includegraphics[width=\linewidth,trim=0cm 0cm 0cm 0cm,clip]{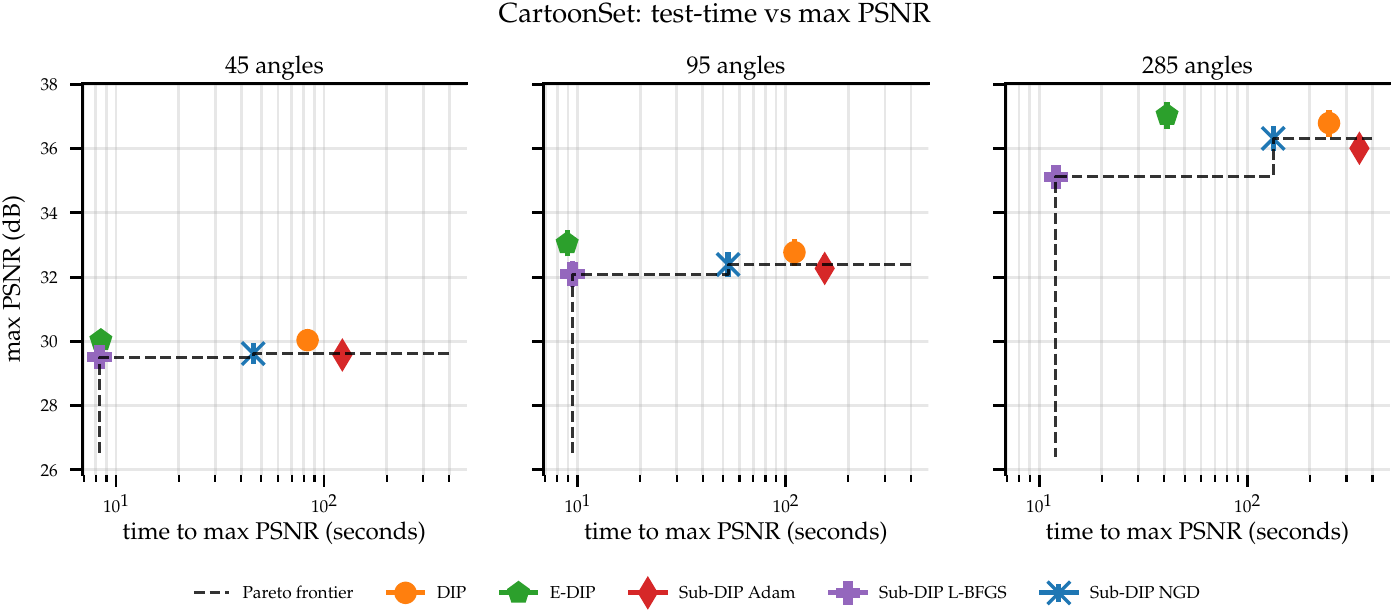}
    \caption{Pareto-curves of max. PSNR vs optimisation time on the CartoonSet for reconstructions from 45 angles (left), 95 angles (middle) and 285 angles (right). We provide mean and standard deviation of the PSNR computed across 50 cartoon images. Note that the x-axis is given in log-scale and that the std is hard to observe due to the shared range along the y-axis.
    }
    \vspace{-0.15cm}
    \label{fig:appendix_cartoon_pareto_max}
\end{figure*}

\begin{figure}[htb]
\vspace{-0.0cm}
    \centering
    \begin{minipage}{0.32258\linewidth}
        \centering
        45 angles\\[0.05cm]\hrule\vspace*{0.1cm}
        \includegraphics[width=\textwidth]{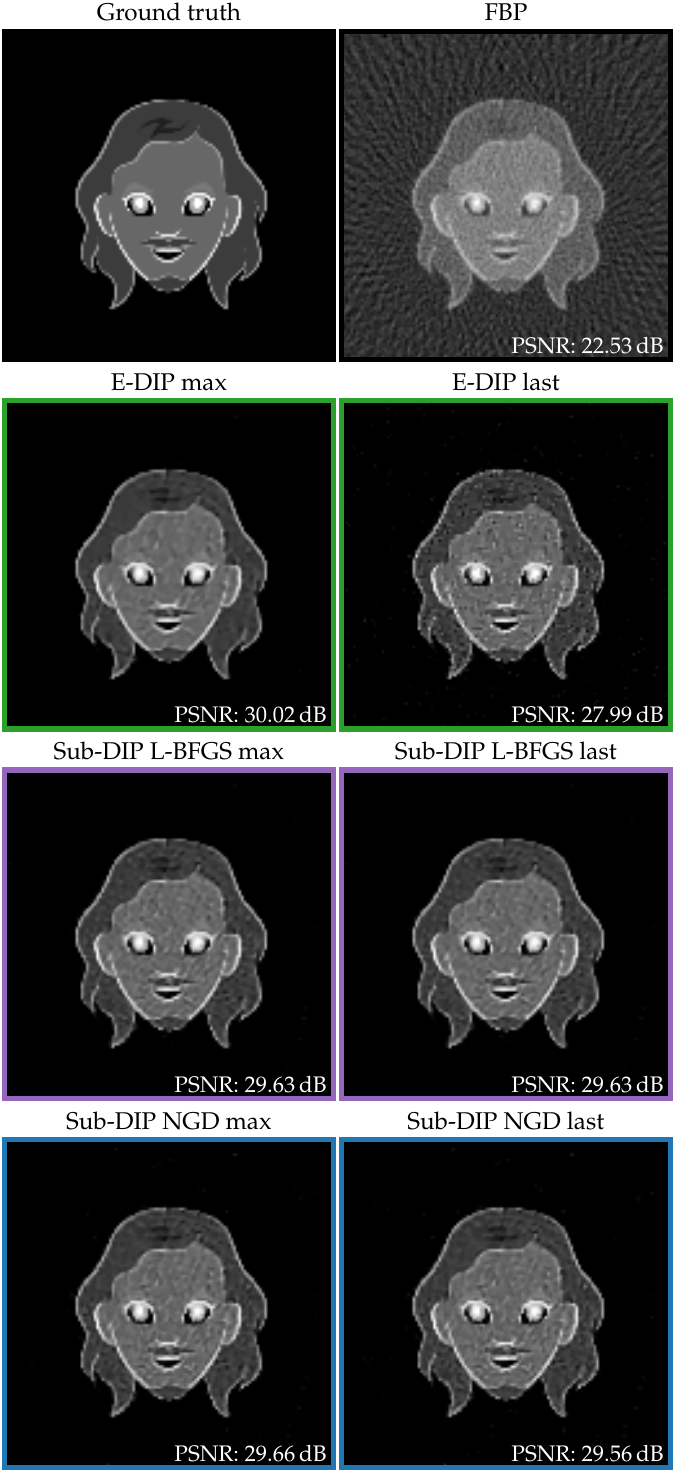}
    \end{minipage}\hfill%
    \begin{minipage}{0.32258\linewidth}
        \centering
        95 angles\\[0.05cm]\hrule\vspace*{0.1cm}
        \includegraphics[width=\textwidth]{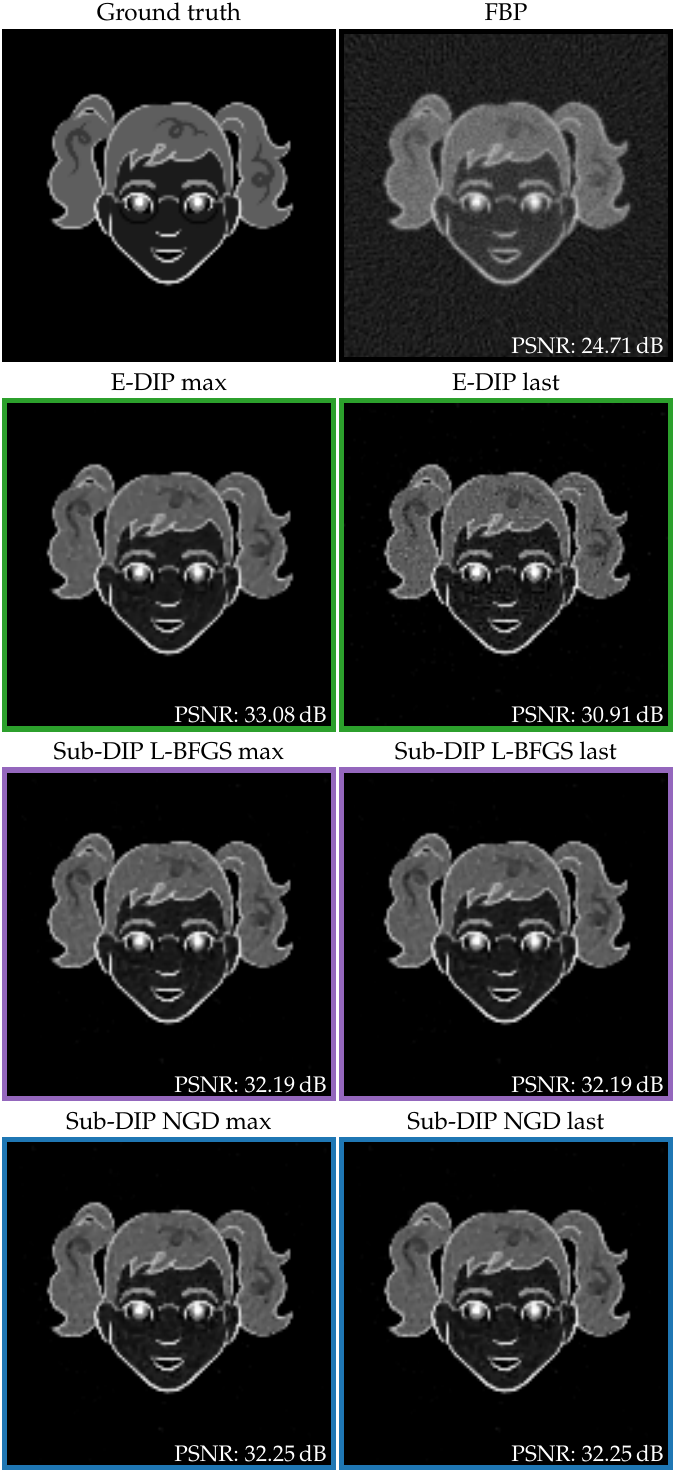}
    \end{minipage}\hfill%
    \begin{minipage}{0.32258\linewidth}
        \centering
        285 angles\\[0.05cm]\hrule\vspace*{0.1cm}
        \includegraphics[width=\textwidth]{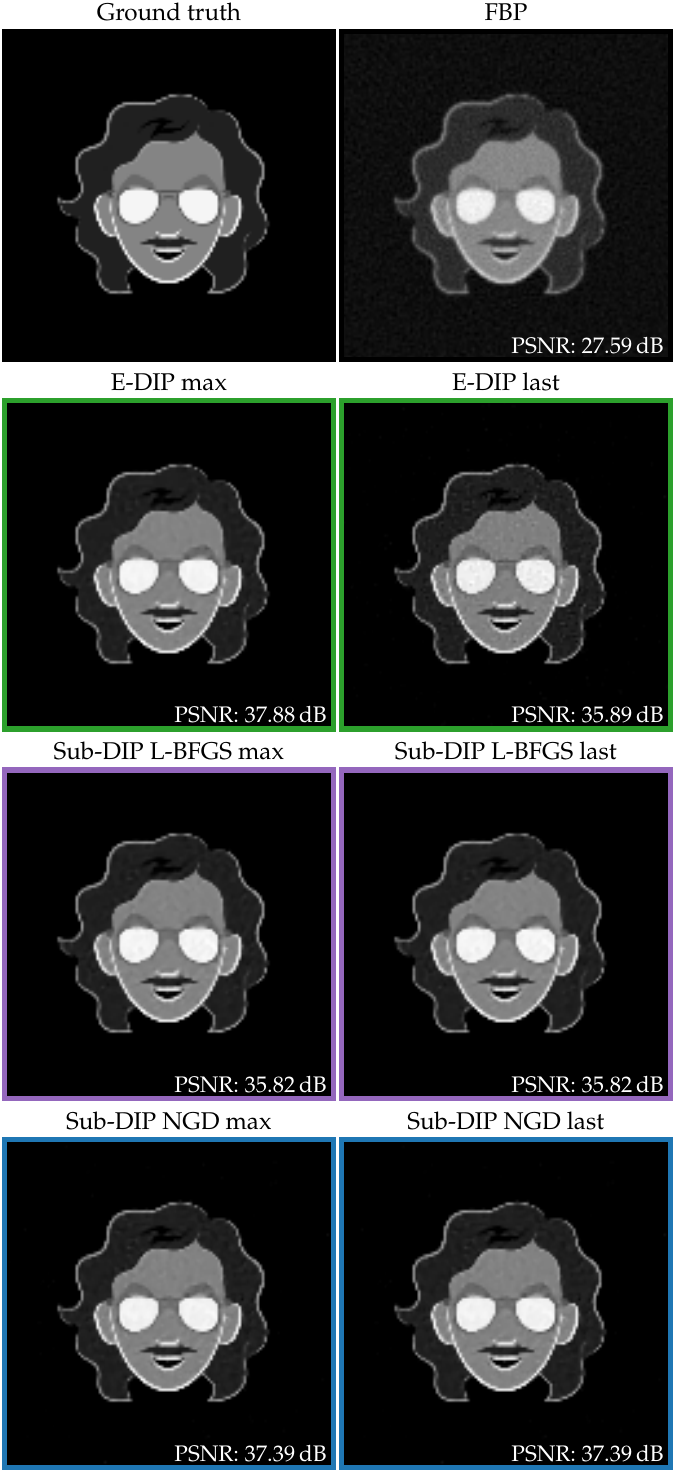}
    \end{minipage}
    \vspace{-0.1cm}
    \caption{Reconstructions for the CartoonSet, from 45, 95 and 285 acquisition angles for different example images and DIP methods.}
    \label{fig:appendix_cartoon_images_45}
    \vspace{-0.1cm}
\end{figure}

In \cref{fig:appendix_cartoon_pareto}, we show Pareto-optimality of subspace methods. Both Sub-DIP NGD and L-BFGS show performances superior to DIP and E-DIP in terms of time to convergence and conv. PSNR, for all the studied levels of problem ill-posedness (that is, with respect to the number of angles). This confirms observations made in \cref{sec:experiments} and complements results in \cref{fig:cartoon_ablation}.
We note that conv. is based on the stopping criterion in \cref{alg:1}. Interestingly, the time required to reach the conv. PSNR values is for all the studied methods largely independent of the number of observation angles. Just as expected, all methods' conv. PSNR is higher for more well-conditioned settings. 
Furthermore, if the time at max PSNR is considered, as in \cref{fig:appendix_cartoon_pareto_max}, then E-DIP is within the Pareto optimality frontier.
In \cref{fig:appendix_cartoon_images_45} we show three examples of reconstructions on the CartoonSet for the three angle settings in the ablative study in \cref{subsec:cartoon}. Even for the sparsest view (45 angles), Sub-DIP reconstructions exhibit barely any noise.

\setlength{\textfloatsep}{15pt}%

\begin{algorithm2e}
    \KwData{metric $g(\cdot)$, patience $\mathfrak{p}$, decrease proportion $\delta$}
    $g_{\rm{min}} \xleftarrow{} \infty$,  $i \xleftarrow{} 0$, $i_{\rm{min}} \xleftarrow{} \infty$ \\
	\While{$i \leq i_{\rm{min}} + \mathfrak{p}$}{
	\If{$g(i) < \delta \cdot g_{\rm{min}} $ }{
	    $g_{\rm{min}} \xleftarrow{}  g(i)$ and 
    $i_{\rm{min}}  \xleftarrow{} i$
	}
	$i \xleftarrow{} i + 1$
	}
\KwResult{$i_{\rm{min}}$}
\caption{Early stopping criterion \label{alg:1}}
\end{algorithm2e}

\paragraph{Selection and construction of subspaces}

\begin{figure*}[htb]
\vspace{-0.05cm}
    \centering
    \includegraphics[width=0.7\linewidth,trim=0cm 0cm 0cm 0cm,clip]{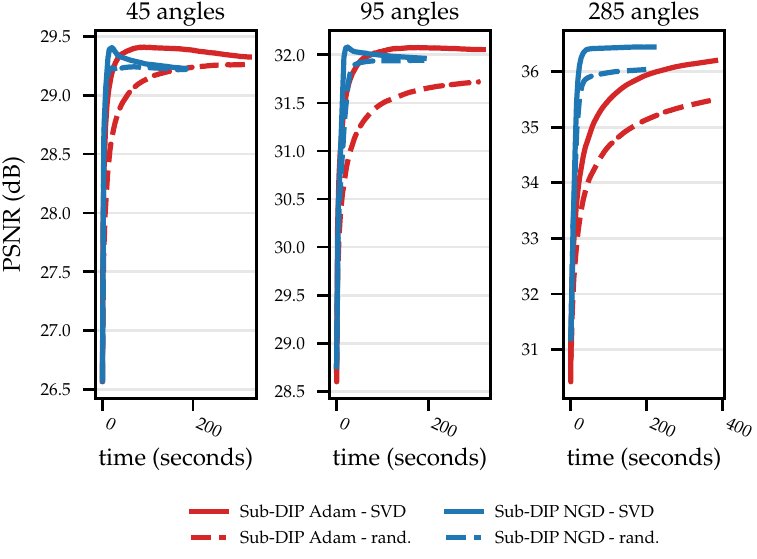}
    \caption{Average PSNR trajectories for 45, 95, and 285 angles, comparing a randomly selected low-dimensional subspace, and a subspace computed through SVD on the pre-training trajectory, on Sub-DIP Adam and Sub-DIP NGD. 
    }
    \vspace{-0.15cm}
    \label{fig:svd_vs_random}
\end{figure*}

Firstly, we compare the reconstruction quality of Sub-DIP NGD utilising an SVD basis extracted from a pretraining trajectory (as detailed in the paper) with Sub-DIP NGD using a randomly sampled unit-norm basis of equal dimensionality.
The goal is to investigate if there is a benefit\cref{fig:svd_vs_random} shows PSNR reconstruction trajectories, averaged over 25 images.
The result confirms that using a subspace based on the pre-training trajectory has a clearly superior performance, justifying the choices made in the paper.
The effect is more pronounced in the less ill-posed problems (that is, as the number of angles increases).

\begin{figure*}[htb]
\vspace{-0.05cm}
    \centering
    \includegraphics[width=0.7\linewidth,trim=0cm 0cm 0cm 0cm,clip]{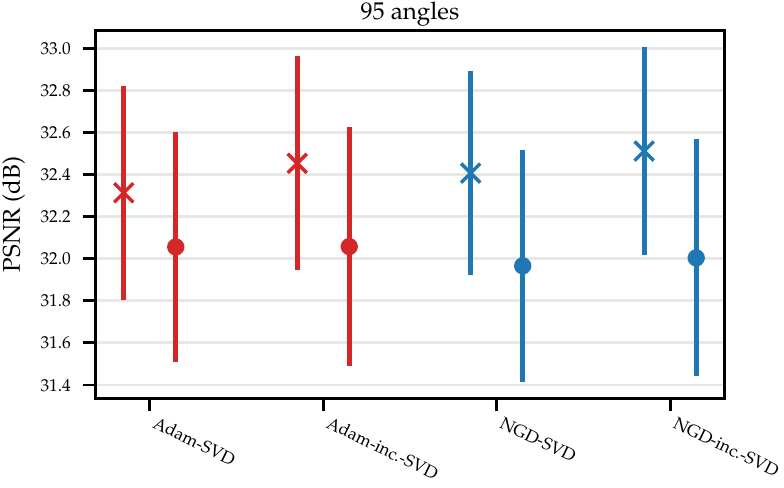}
    \caption{Comparison of max and conv. PSNR of traditional and incremental SVD approaches, used to construct the pre-training subspace, for Sub-DIP Adam and Sub-DIP NGD.
    }
    \vspace{-0.15cm}
    \label{fig:svd_vs_incsvd}
\end{figure*}

In \cref{fig:svd_vs_incsvd}, we compare the reconstruction quality with respect to the numerical scheme that is used to compute the utilised low-dimensional SVD space. 
Specifically, our comparison involves the traditional SVD method, which entails explicitly constructing a matrix of parameters sampled at various points along the training trajectory, followed by computing its SVD. We contrast this with the more computationally efficient approach known as incremental SVD \cite{brand2002incremental}.
The results for max and conv. PSNR (averaged over 25 images) report that the two methods show on par performance.

\subsection{Additional \textmu CT Walnut results and discussion}

\begin{figure*}[t!]
    \centering
    \includegraphics[width=\linewidth]{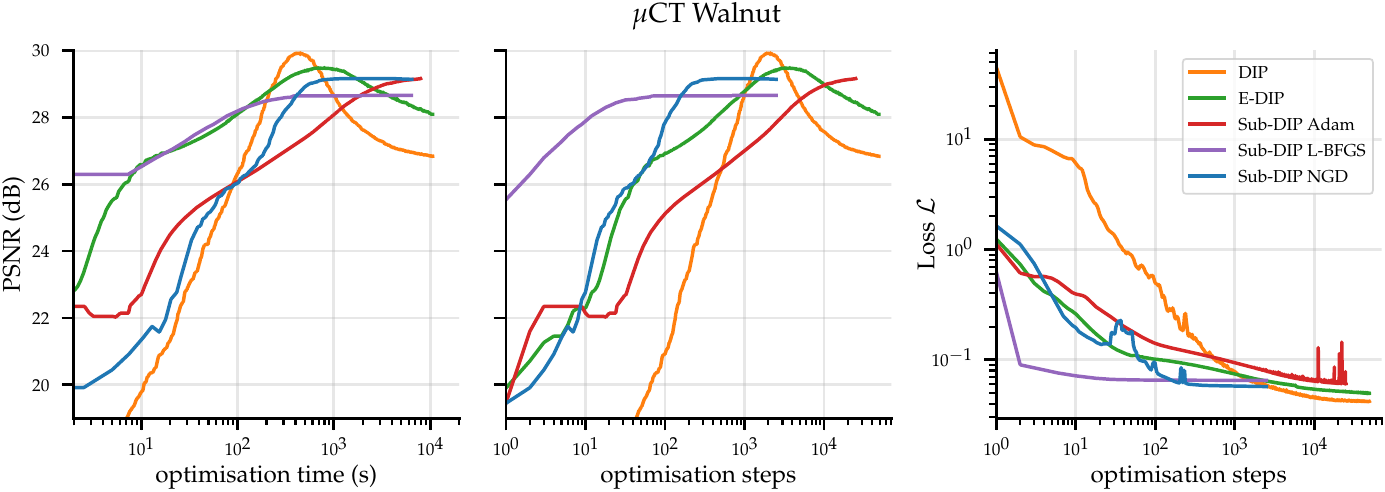}
    \caption{The training curves of DIP, E-DIP and three versions of Sub-DIP on the \textmu CT Walnut dataset. PSNR vs time (left), PSNR vs steps (middle) and loss vs steps (right). All curves (expect for E-DIP) are averaged over 3 seeds, which affect the initialisation point for the U-Nets, initialisation of the subspace parameters and random probes used for NGD. Note that the x-axis is given in log-scale.
    }
    \label{fig:appendix_walnut_curves}
\end{figure*}

\paragraph{Optimisation trajectories} \cref{fig:appendix_walnut_curves} shows optimisation trajectories of all the used optimisers on the Walnut dataset, cf. \cref{subsec:walnut}. We study the optimisation behaviour in terms of PSNR vs time, PSNR vs steps and loss vs steps. As discussed in the main text, second order subspace methods converge in less time than first order method. 
The rightmost plot shows that the loss functions for DIP and E-DIP decrease at a constant rate in the log of the number of steps, even after passing their PSNR peak. In contrast, loss curves of subspace methods saturate at their minimum values (at around $7e-2$), which coincides with their max PSNR.

\subsection{Additional Mayo results and discussion}

\begin{figure*}
\vspace{-0.05cm}
    \centering
    \includegraphics[width=\linewidth]{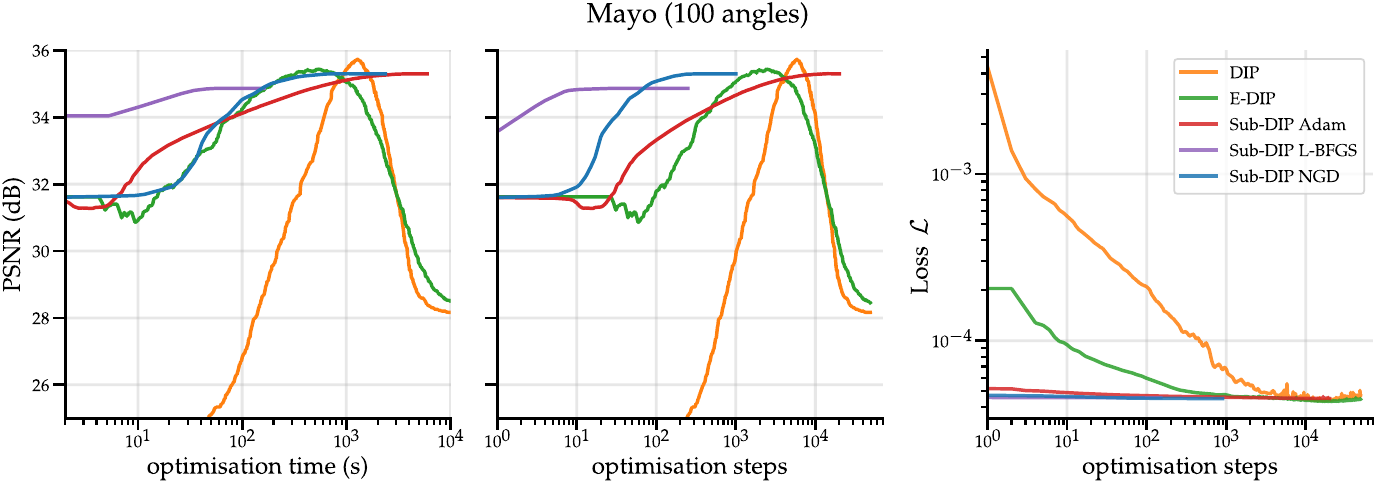}
    \caption{The optimisation curves of DIP, E-DIP and three versions of Sub-DIP on the 100 angle Mayo dataset. PSNR vs time (left), PSNR vs steps (middle) and loss vs steps (right). All curves are averaged over 10 images. Note that the x-axis is given in log-scale.  
    }
    \label{fig:appendix_mayo_100_convergence}
\end{figure*}

\begin{figure*}
\vspace{-0.05cm}
    \centering
    \includegraphics[width=\linewidth]{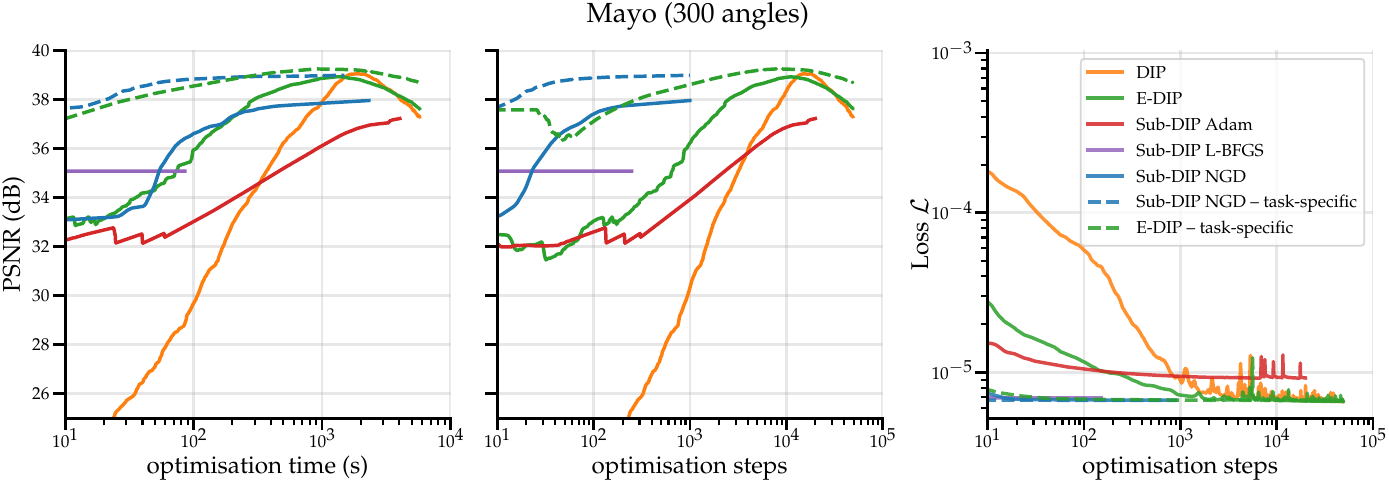}
    \caption{The optimisation curves of DIP, E-DIP and three versions of Sub-DIP on the 300 angle Mayo dataset. PSNR vs time (left), PSNR vs steps (middle) and loss vs steps (right). All curves are averaged over 10 images. Note that the x-axis is given in log-scale.
    }
    \label{fig:appendix_mayo_300_convergence}
\end{figure*}

\paragraph{Optimisation trajectories} 
Figs. \ref{fig:appendix_mayo_100_convergence} and \ref{fig:appendix_mayo_300_convergence} show the optimisation trajectories for the Mayo Clinic dataset on 100 and 300 angle CT tasks, respectively. 
Figures on the left study the PSNR with respect to elapsed time; figures in the middle study PSNR vs number of optimisation steps, and figures on the right study optimisation loss vs number of optimisation steps. 
The results consistently show that second order subspace methods exhibit fast and stable convergence, with no observable performance degradation. 
On the other hand, DIP and E-DIP show high max PSNR, but subsequently overfit to noise, as expected. Moreover, in \cref{fig:appendix_mayo_300_convergence}, we compare the performance of Sub-DIP NGD and E-DIP, where $\theta_{\rm pre}$ and $U$ are obtained using a dataset of images of a similar distribution and structure to the Mayo dataset.

Namely, dashed lines indicate optimisation trajectories with the initial parameters and extracted basis selected through pre-training on the LoDoPaB dataset \cite{leuschner2021lodopab}.
The results indicate that this, task-specific, pre-training allows faster and overall improved performance behaviour. 

\section{Description of our implementation of Natural Gradient Descent}\label{app:NGD}

The standard NGD iteration \citep{Amari2013information}, for a smooth function $\mathcal{L}_\gamma$, is given by
\begin{align}\label{eqn:ngd}
c_{t+1} = c_t - \alpha_t \tilde F(c_t)^{-1} \nabla \mathcal{L}_\gamma(c_t),
\end{align}
where $\nabla\mathcal{L}_\gamma$ is the gradient of the loss function (including the TV regulariser's contribution), $\alpha_t>0$ is a step-size and $\tilde F(c_t)$ is the {exact Fisher information matrix (FIM) computed at $c_t$}. Ignoring the regulariser's contribution, FIM is of the form
\begin{gather}\label{eqn:F_appx}
\tilde F(c) = (A J_f M U)^\top A J_f M U.
\end{gather}
We derive the FIM with respect to $\theta$ from its definition as the expected outer-product between gradients of the projection error term, see \cref{eq:dip_tv}.
The gradient of the data-fitting term with respect to $\theta$ is
\begin{align}\label{eqn:gradlik}
    \nabla_{\theta} \left [ \tfrac{1}{2} \|Af(x^{\dagger}, \theta) - y\|^{2}_{2} \right ] &= \nabla_{x}\left[\tfrac{1}{2} \|Ax - y\|^{2}_{2} \right]\big\vert_{x=f(x^{\dagger}, \theta)}\!\!\nabla_\theta f(x^{\dagger}, \theta)\nonumber\\
    &=(Af(x^{\dagger}, \theta)-y)^\top AJ_f,
\end{align}
with $J_f=\nabla_\theta f(x^\dagger,\theta)$.
By definition, we then have
\begin{align}\label{eqn:F_derv}
    \tilde F (\theta)%
    &= \mathbb{E}_{v \sim \mathcal{N}(\mu, I_{d_{y}})} \!\left [(J_{f}^{\top}A^{\top}(Af(x^{\dagger}, \theta) - v))^{\otimes^{2}} \right ]  \nonumber
    \!=\!\mathbb{E}_{v \sim \mathcal{N}(\mu, I_{d_{y}})} \left [
    J_{f}^{\top}A^{\top} (Af(x^{\dagger}, \theta) - v)^{\otimes^{2}}AJ_{f}\right] \nonumber\\&=J_{f}^{\top}A^{\top} \mathbb{E}_{v \sim \mathcal{N}(\mu, I_{d_{y}})} \left [(Af(x^{\dagger}, \theta) - v)^{\otimes^{2}}\right] AJ_{f},
\end{align}
where $\mu$ is defined as $Af(x^{\dagger}, \theta)$ and $\otimes^2$ denotes the outer product of a vector $v$ with itself (i.e., $z^{\otimes^2} = zz^{\top}$). The expectation in \cref{eqn:F_derv} simplifies to 
\begin{align*}
    \mathbb{E}_{v \sim \mathcal{N}(\mu, I_{d_{y}})} \!\!\left [(Af(x^{\dagger}, \theta) \!-\! v )^{\otimes^{2}}\right] &= \!\mathbb{E}_{v \sim \mathcal{N}(\mu, I_{d_{y}})} \!\left [Af(x^{\dagger}, \theta)f(x^{\dagger}, \theta)^{\top}A^{\top} \!+ \!vv^{\top} \!-\! 2vf(x^{\dagger}, \theta)^{\top}A^{\top}\right]  \\ 
    &=Af(x^{\dagger}, \theta)f(x^{\dagger}, \theta)^{\top}A^{\top} + \mathbb{E}_{v \sim \mathcal{N}(\mu, I_{d_{y}})} [vv^{\top}] \\ & \quad\quad\quad - 2 \mathbb{E}_{v \sim \mathcal{N}(\mu, I_{d_{y}})}[v]f(x^{\dagger}, \theta)^{\top}A^{\top}  \\ 
    &= I_{d_y}. 
\end{align*}
Thus, the FIM with respect to $\theta$ is given by
\begin{equation}\label{eqn:F_appx_theta}
    \tilde F (\theta) =J_{f}^{\top}A^{\top}AJ_{f}.
\end{equation}

Alternatively, the equivalence between NGD and generalised Gauss-Newton (GGN) methods \cite{Kunstner2019Limitations, schraudolph2002fast}, valid for exponential family likelihoods \citep{Kunstner2019Limitations}, can be exploited for this problem. Namely, the data fidelity is proportional to the negative exponential log-likelihood under the noise model in \cref{eq:inverse_problem}. The Hessian $H$ of the data fidelity with respect to $\theta$ is given by
\begin{align}\label{eqn:Hess}
    \nabla^{2}_{\theta}\| Af(x^{\dagger}, \theta) - y \|^{2}_{2} &=
    H_{f}^{\top}A^{\top}Af(x^{\dagger}, \theta) - H_{f}^{\top}A^{\top}y + J_{f}^{\top}A^{\top}AJ_{f},
\end{align}
where $H_{f} := \nabla^{2}_{\theta} f(x^\dagger, \theta)\in \mathbb{R}^{d_{\theta} \times d_{\theta}}$ is the network Hessian. GGN methods are then recovered by ignoring the second order terms in \cref{eqn:Hess}, giving
\begin{equation}\label{eqn:ggn}
    G(\theta) = J_{f}^{\top}A^{\top}AJ_{f}= \tilde F(\theta).
\end{equation}
Note that \cref{eqn:F_appx} is then trivially recovered from \cref{eqn:F_appx_theta} or \cref{eqn:ggn} by introducing the network reparametrisation in \cref{eq:objective_reparametrisation}. Namely, an analogous computation yields
\begin{align}\label{eqn:gradlik_coeffs}
    \nabla_{c} \left [ \tfrac{1}{2} \|Af(x^{\dagger}, \gamma(c)) - y\|^{2}_{2} \right ] =(Af(x^{\dagger}, \gamma(c))-y)^\top AJ_fMU.
\end{align}
Plugging this in and taking the expectation recovers \cref{eqn:F_appx}.

Assuming the NGD-GGN equivalence, the curvature of the DIP loss \eqref{eq:dip_tv} has no contribution coming from the TV regulariser, since the latter consists of the absolute value of the finite differences of pixel values (cf. \eqref{eq:TV_equation}) and thus almost everywhere has zero second derivatives.
Note also that for image restoration tasks in \cref{sec:natural_images}, TV regularisation is not utilised.

We depart from \cref{eqn:ngd} in two ways. First, we use a stochastic estimate of the Fisher and second, we relax the update rule by adding a number of hyperparameters. These are set adaptively with a modified version of \citep{Martens2015kron}'s Levenberg–Marquardt-style algorithm, described below.

\subsection{Stochastic update of the Fisher information matrix}

As indicated in \cref{subsec:2O_optimisation}, we compute a Monte-Carlo estimate of the FIM at step $t$ as 
\begin{gather}\label{eqn:fisher_probing_appendix}
\hat{F}_{t} = \frac{1}{n} \sum_{i=1}^n (z_i^\top A J_f M U)^\top z_i^\top A J_f M U ,
\end{gather}
with $J_f \coloneqq \nabla_{\theta} f(x^{\dagger}, \gamma(c_t)) \in \R^{d_{x} \times \dtheta }$ being the Jacobian of the U-Net at the current full-dimensional parameter vector, and $n$ the number of random probes $z_i \sim \mathcal{N}(0, I_{d_{y}})$. 
This aims at overcoming the computational intractability arising from $J_{f}$.  That is, we approximate the matrix–matrix multiplications in \cref{eqn:ngd} via Monte-Carlo sampling \cite{MartinssonTropp:2020}. Then we update the FIM moving average as
\begin{align}\label{eq:FMA_appendix}
    F_{t+1} &= \beta F_t + (1 - \beta) \hat{F}_{t} \quad \textrm{and} \quad \beta \in (0,1).
\end{align}
We use our \emph{online FIM estimate $F_t$} to estimate the descent direction according to the natural gradient at $c_t$ as $\Delta_t = -F_{t}^{-1} \nabla_c \mathcal{L}_\gamma(c_t)$.

To evaluate \eqref{eqn:fisher_probing_appendix} we use $n=100$ probes per optimisation step for the CartoonSet ablative study in \cref{subsec:cartoon}. On the Walnut and Mayo datasets (in \cref{subsec:walnut} and \cref{subsec:mayo}), due to the increased computational cost of the Jacobian vector products,  we only use $n=50$ probes per optimisation step. 
Finally, across all experiments, to update the FIM moving average, cf. \eqref{eq:FMA_appendix}, $\beta$ is kept fixed to 0.95. 

\subsection{Adaptively fine-tuning the FIM hyperparameters}

To compute a new parameter setting $c_{t} + \delta$, we choose $\delta$ to locally minimise a quadratic model of the training objective $\mathcal{L}_\gamma$, defined as
\begin{align}\label{eqn:ngd_quadmodel} 
M_t(\delta)=\mathcal{L}_\gamma(c_t) + \nabla_c \mathcal{L}_\gamma(c_t)^\top\delta + \dfrac{s}{2}\delta^\top ( \lambda I_{\dsub} + \tilde  F(c_t) )\delta.
\end{align}
Since $\tilde F(c_t)$ is the FIM (guaranteed PSD) and not the Hessian, the above can be seen as a convex approximation of the second order Taylor series expansion of $\mathcal{L}_\gamma$ at $c_t$. 
Since neural network loss-functions are non-quadratic and non-convex, 
the FIM may provide a poor approximation to the loss. To correct for this, we introduce two parameters, $s$ and $\lambda$, leading to the modified curvature $s(\lambda I_{\dsub} + \tilde F)$. The parameter $\lambda>0$ ensures the positive definiteness of the FIM that may be violated due to numerical instabilities, and also provides an isotropic increase in curvature \cite{Martens2022review}, limiting the norm of the loss gradient and bringing it closer to the steepest descent direction.
Novel to our method is the scaling parameter $s \in (0, 1)$, which can reduce the effect of the curvature on the quadratic model, thus allowing larger stepsizes.

We employ the common Levenberg-Marquardt style methodology (see \cite[Section 8.5]{Martens2012free})  to update both the damping parameter $\lambda$ and the scaling parameter $s$. This involves computing the ratio
\begin{align}\label{eqn:ngd_rho} 
    \rho=\frac{\mathcal{L}_\gamma(c_t+\Delta_t) - \mathcal{L}_\gamma(c_t)}{M_t(\Delta_t)-M_t(0)}.
\end{align}
Thus, for $\rho$ close to $1$ the quadratic model is good at approximating the objective, and if $\rho$  is substantially smaller than $1$ then it is a poor estimator. Following \citep{Martens2012free}, $\rho$ is evaluated every $T=5$ iterations. If $\rho<0.25$ the damping is updated via $\lambda\leftarrow \left(\frac{3}{4}\right)^{-T}\lambda$. Conversely, if $\rho>0.75$ then $\lambda\leftarrow \left(\frac{3}{4}\right)^{T}\lambda$.
If needed, the resulting value is clipped to ensure it stays in the interval $[\lambda_{\text{min}}, 100]$.

Across all experiments, the damping coefficient $\lambda$ is initialised to 100. The different dimensionality of the subspace $\dsub$ necessitates adjusting $\lambda_{\text{min}}$ in order to avoid numerical instabilities when solving the linear system against $F(c_t)$, required to compute the update direction $\Delta_{t}$. 
Due to the small dimensionality of $\dsub$ used in the CartoonSet ablative study, $\lambda_{\text{min}}$ is set to $10^{-8}$.
On the Walnut and the Mayo data, we set $\lambda_{\text{min}}$ to $1$ due to the high-dimensionality of the considered subspace.

Ideally, the parameter $s$ would be small during the early iterations and would increase towards $1$ as we approach the optimum, where we want optimisation to slow down. We use a similar update condition: if $\rho<0.95$ then $s \leftarrow\left(\frac{3}{4}\right)^{-T}s$ and if $\rho>1.05$ then $s\leftarrow\left(\frac{3}{4}\right)^{T}s$.
If needed, the resulting value is clipped to ensure it stays in the interval $[s_{\text{min}},1]$. Note that the rule under which $s$ is updated is much tighter that the one used for the damping parameter $\lambda$. This is done to ensure larger step-sizes can be taken in the early stages of the optimisation, speeding up the convergence.

We set $s_{\text{min}}$ to $10^{-3}$ for the CartoonSet (across all three angles setting). For the Walnut and the Mayo datasets, as well as for the image restoration tasks, we set $s_{\text{min}}$ to $5\times 10^{-6}$.

\subsection{Getting the final parameter update }

We further speed up the convergence by introducing a momentum update, as in \citep{scarpetta1999matrix, Martens2015kron}. 
This results in update directions of the form $\delta=\alpha_t\Delta_t + \mu_t\delta_0$, 
where $\Delta_t = -F_{t}^{-1} \nabla_c \mathcal{L}_\gamma(c_t)$, with $F_t$ being our moving average estimate of the FIM, and $\delta_0$ is the direction of the previous update. 
Coefficients are then updated as $c_{t+1}=c_t+\delta$.
Parameters $\alpha_t$ and $\mu_t$ are chosen by minimising  the local quadratic model $M_t$. Plugging such a $\delta$ into $M_t$ and minimising over $\alpha_t$ and $\mu_t$ gives a two-dimensional linear system
\begin{align}\label{eqn:ngd_momentum}
\begin{pmatrix} \alpha_t\\ \mu_t \end{pmatrix} = - s^{-1} \begin{pmatrix}\Delta_t^\top\tilde F(c_t)\Delta_t + \lambda\|\Delta_t\|_2^2 & \Delta_t\Tilde F(c_t)\delta_0 +\lambda\Delta_t^\top\delta_0\\\Delta_t^\top\tilde F(c_t)\delta_0 + \lambda\Delta_t^\top\delta_0 & \delta_0^\top\Tilde F(c_t)\delta_0 +\lambda\|\delta_0\|^2_2\end{pmatrix}^{-1}\begin{pmatrix}\nabla\mathcal{L}_\gamma(c_t)^\top\Delta_t\\ \nabla\mathcal{L}_\gamma(c_t)^\top\delta_0\end{pmatrix}.
\end{align}
Note that, although it is not tractable to compute the full FIM at every optimisation step and we use a rolling estimate $F_t$, we may interact with the true FIM $\tilde F$ through matrix vector products. This allows solving the above systems quickly.

\section{Additional experimental setup description}\label{app:experimental_setup}

\subsection{Raw PSNR vs min-loss PSNR}

\begin{figure*}[htb]
\vspace{-0.05cm}
    \centering
    \includegraphics[width=\linewidth]{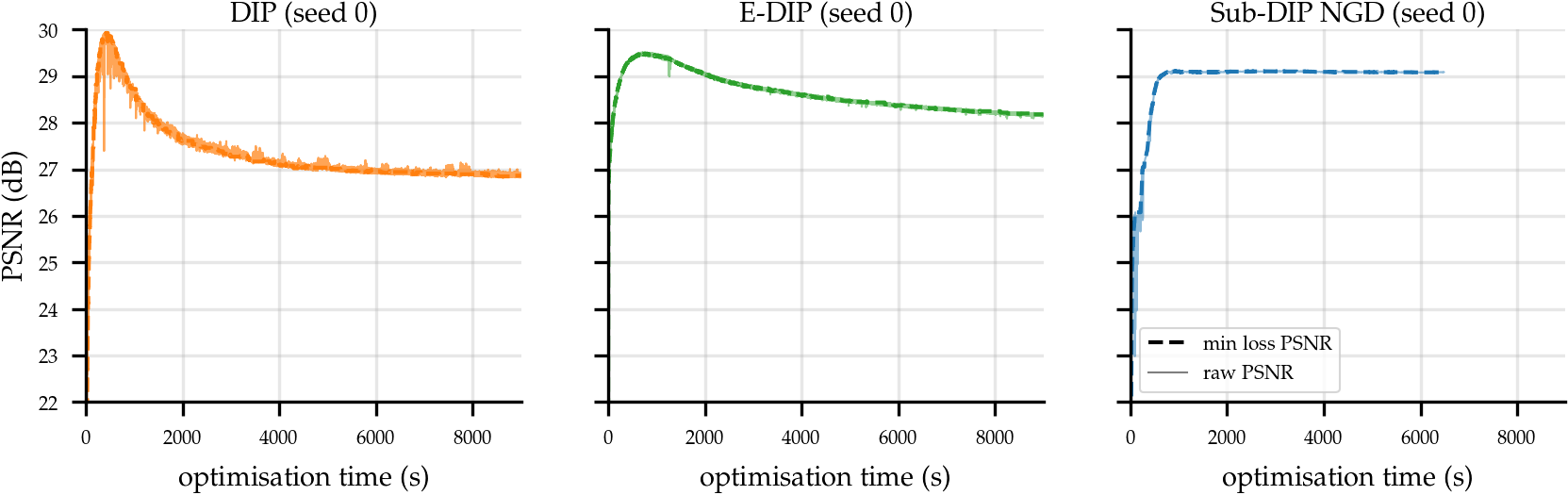}
    \caption{Min-loss and raw PSNR values obtained when reconstructing the Walnut data described in \cref{subsec:walnut} using seed 0. Min-loss PSNR acts as a smoothed version of raw PSNR, avoiding jumps from optimisation instabilities (left).
    }
    \vspace{-0.15cm}
    \label{fig:appendix_walnut_min_vs_psnr}
\end{figure*}

All the reported PSNR values are obtained using the min-loss PSNR strategy standard in the DIP literature \citep{baguer2020diptv}. For sparse problems, both the training loss and ``raw'' reconstruction PSNR can exhibit very rapidly varying behaviour across optimisation steps. In order to display PSNR values, at each optimisation step, we define the min-loss PSNR as the PSNR corresponding to the time-step with lowest training loss up to the current time. We illustrate the difference between raw and min-loss PSNR in Fig. \ref{fig:appendix_walnut_min_vs_psnr}. Interestingly Sub-DIP NGD differs from full parameter methods in that it does not suffer from noisy optimisation, showing that the approach enjoys excellent stability during the training.

\subsection{CT forward model operators}

\paragraph{Parallel-beam geometry for CartoonSet}
The parallel-beam geometries described in \cref{subsec:cartoon} are computed using the ODL library \cite{adler2017operator} 
({https://github.com/odlgroup/odl}) with the ``\texttt{astra\_cuda}'' backend \cite{aarle2015astra}.
The angles are adjusted to start with $0$ instead of the default $\pi/(2\cdot\text{num\_angles})$.
The number of detector pixels is chosen automatically by ODL such that the discretised image is sufficiently sampled.
Since the back-projection operation would approximate the adjoint of the forward projection (due to discretisation differences), we assemble the matrix by calling the forward projection operation for every standard basis vector, $A = A [e_{1}, e_{2},\,...\, e_{d_x}]$, where $d_x=128^2$ is the total number of image pixels.
The resulting matrix $A$ and its transposed $A^\top$ are used for the forward model and its adjoint, respectively.

\paragraph{Pseudo-2D fan-beam geometry for Walnut}
We restrict the 3D cone-beam ASTRA geometry provided with the dataset to a central 2D slice for the first Walnut at the second source position (tubeV2). To do so, we use a sub-sampled set of measurements, which corresponds to a sparse fan-beam-like geometry. From the original 1200 projections (equally distributed over $2\pi$) of size $972\times 768$ we first select the appropriate detector row matching the slice position (which varies for different detector columns and angles due to a tilt in the setup), yielding measurement data of size $1200\times 768$. We then sub-sample in both angle and column dimensions by factors of $20$ and $6$, respectively, leaving $d_y = 60\times 128 = 7680$ measurements.
As for the operator used in the ablation study on the CartoonSet, we assemble the matrix by calling the forward projection operation for every standard basis vector and use $A$ and $A^\top$ for the forward model and its adjoint, but stored in the sparse matrix form because of the large dimensions.
Due to the special selection of detector pixels in order to create the single-slice pseudo-2D geometry, back-projection via ASTRA is not applicable here, so our slightly slower sparse-matrix-based implementation is mandatory.

\paragraph{Fan-beam geometry for Mayo}
We create the fan-beam geometries using ODL with the ``\texttt{astra\_cuda}'' backend.
Source and detector radius are chosen to correspond to 700 image pixels, roughly corresponding to the size of a clinical CT gantry (diameter ca.\ 80\,cm).
The original image size $(512\,\text{px})^2$ is used for the 100 angle case, but we crop an image of size $(362\,\text{px})^2$ for the 300 angle case, thereby restricting the area to the region inside a circle (outside of which some CT images have invalid values) defining a circular field of view.
Like for the CartoonSet, we adjust the angles to start with $0$.
The number of detector pixels is chosen automatically by ODL such that the discretised image is sufficiently sampled.
We assemble matrices $A$ and $A^\top$ as for CartoonSet and Walnut datasets.

\subsection{Architectures}\label{app:arch}
\begin{figure*}[htb]
    \centering
    \includegraphics[width=0.75\textwidth]{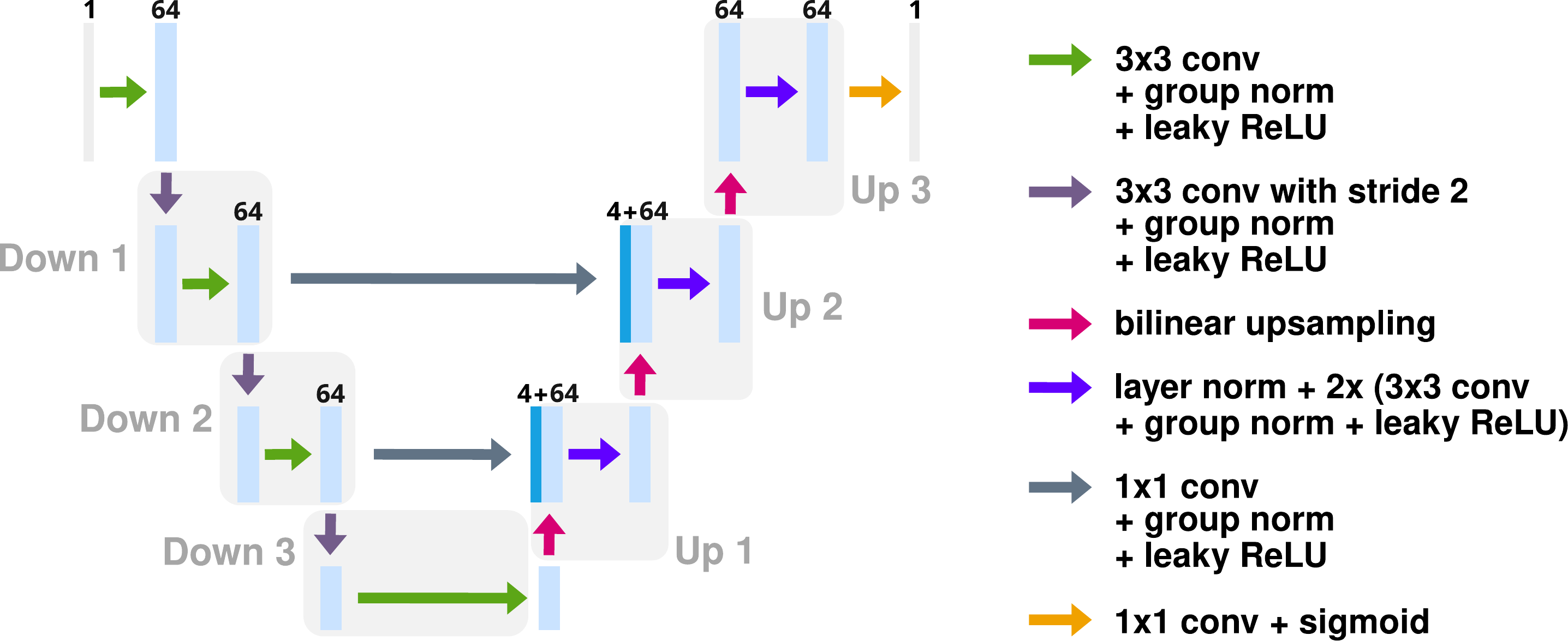}
    \caption{U-Net architecture used with the CartoonSet data. The numbers of channels are indicated above each feature vector.}
    \label{fig:architecture_cartoon}
\end{figure*}

\begin{figure*}[htb]
    \centering
    \includegraphics[width=0.75\textwidth]{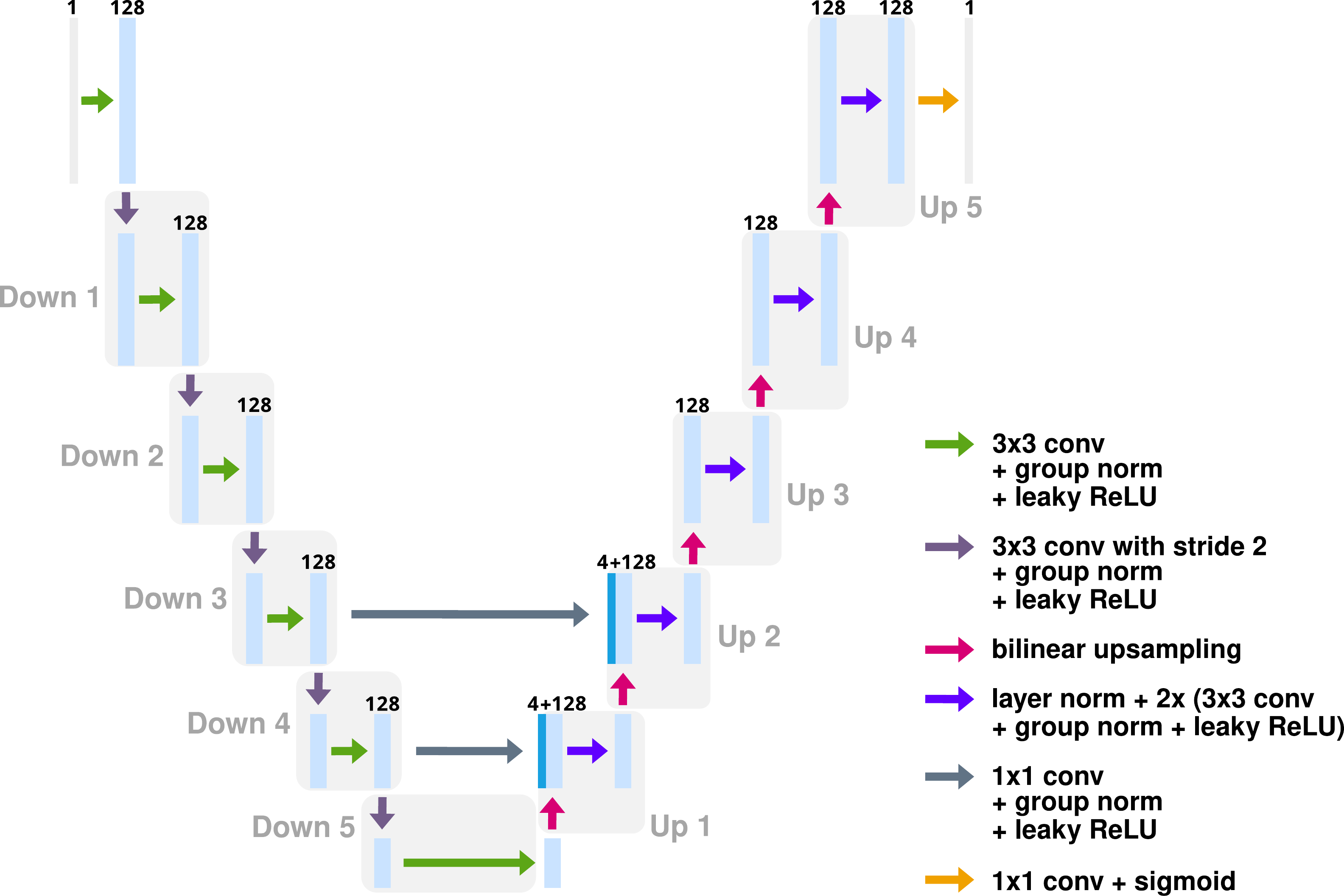}
    \caption{The larger U-Net architecture used with Walnut and Mayo Clinic datasets. The numbers of channels are indicated above each feature vector.}
    \label{fig:unet_arq}
\end{figure*}

\begin{figure*}[htb]
    \centering
    \includegraphics[width=0.75\textwidth]{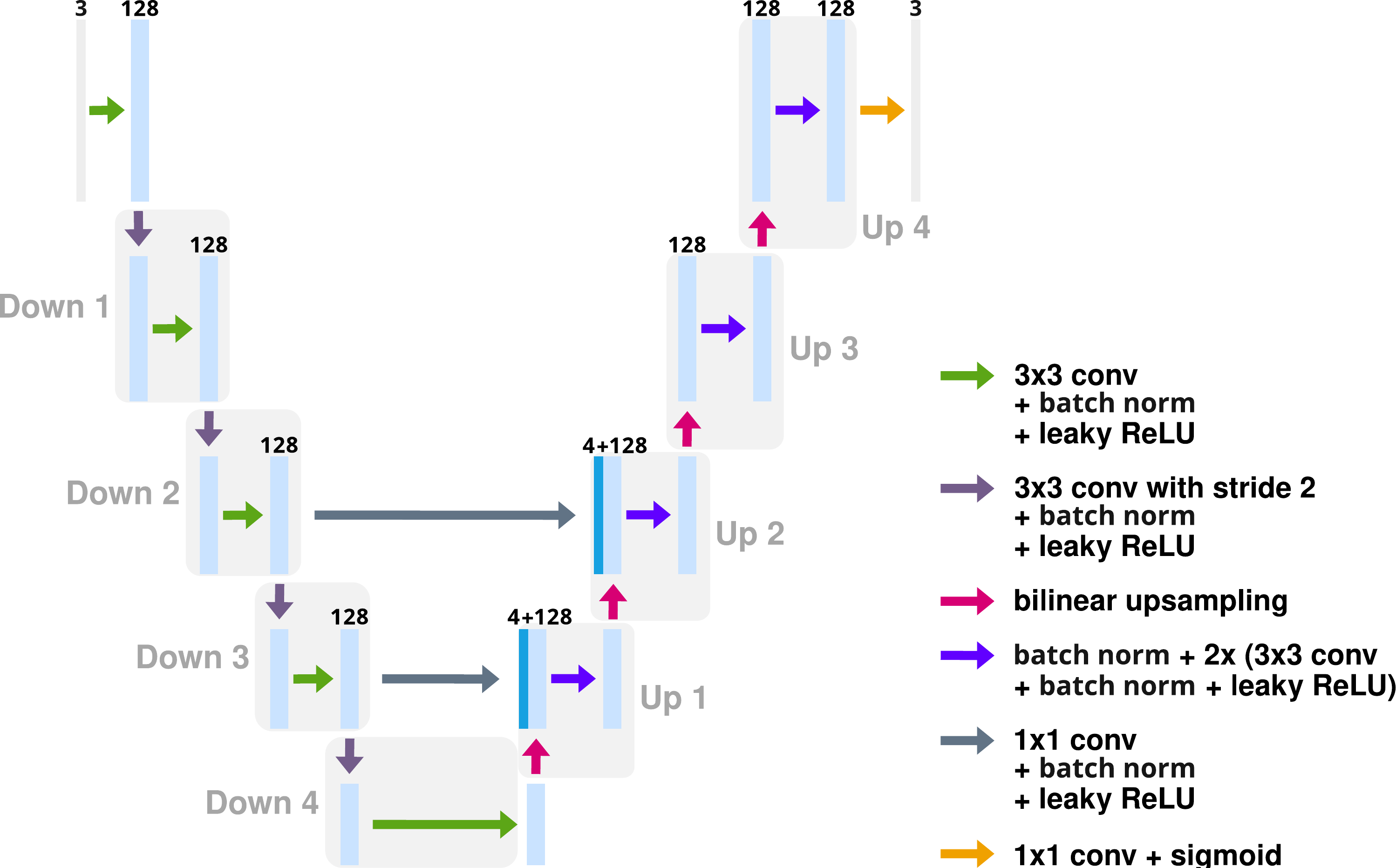}
    \caption{The U-Net architecture used with Set5 dataset of natural images. The numbers of channels are indicated above each feature vector.}
    \label{fig:natural_arq}
\end{figure*}

The U-Net architecture used for the CartoonSet experiments is shown in \cref{fig:architecture_cartoon}. The other tomographic experiments (on the \textmu CT Walnut and Mayo data) use a larger architecture with two more scales and 128 channels in each layer, as shown in \cref{fig:unet_arq}. The architecture used for image restoration experiments is instead shown in \cref{fig:natural_arq}.

\subsection{Other experimental hyperparameters}

\Cref{tab:tv_value_appendix} reports the parameter $\lambda$, used to weigh the contribution of the TV regulariser in \cref{eq:dip_tv}.  
For the CartoonSet, $\lambda$ is selected on a validation set consisting of 5 sample images. Similarly, for the Mayo data (for both 100 and 300 angle settings), $\lambda$ is selected on a validation set of of 3 sample images. For the Walnut dataset, $\lambda$ is selected by visual inspection of the reconstruction.
Across all tomographic experiments, when optimising DIP or E-DIP with Adam, we keep the learning rate to $10^{-4}$ and $3\times 10^{-5}$, respectively. 
Finally, Sub-DIP Adam uses a learning rate of $10^{-3}$.
For image restoration, $\lambda = 0$ accross all settings. 

\begin{table}[H]
\centering
\caption{$\lambda$ values used for TV scaling in \cref{eq:dip_tv}. 
}\label{tab:tv_value_appendix}
\begin{tabular}{ccccccc}
         & \multicolumn{3}{c}{CartoonSet} & Walnut & \multicolumn{2}{c}{Mayo Clinic}\\
         \hline
$\#$ angles  & 45       & 95       & 285      & 120    & 100            & 300 \\
\hline 
$\lambda$ & $3\times 10^{-5}$ & $3\times 10^{-5}$ & $3\times 10^{-5}$ & $6.5 \times 10^{-6}$ & $10^{-4}$ & $10^{-4}$\\
\hline
\end{tabular}
\end{table}

\nocite{Barbano2022Design} \nocite{antoran2022adapting}
\nocite{antoran2022Tomography}
\nocite{antoran2019disentangling}
\end{document}